\documentclass[journal]{IEEEtran}
\usepackage{graphicx}
\usepackage{caption}
\graphicspath{{images/} }
\usepackage{amsmath}
\usepackage{amsthm}
\usepackage{cases}
\usepackage[english]{babel}
\usepackage{algorithm}
\usepackage{algorithmicx}

\usepackage{multirow}
\usepackage{array}
\usepackage{bm}
\usepackage{algorithmicx}
\usepackage{verbatim}
\usepackage{amsfonts,amssymb}

\usepackage{array}
\usepackage{color}
\usepackage{booktabs}
\usepackage{float} 
\usepackage{subfigure}
\usepackage{algpseudocode}

\makeatletter

\newcommand{\Rmnum}[1]{\expandafter\@slowromancap\romannumeral #1@}

\makeatother

\ifCLASSINFOpdf
\else

\fi
\DeclareCaptionLabelFormat{fig}{Fig.~#2}
\begin{document}
\setlength{\abovedisplayskip}{2pt}
\setlength{\belowdisplayskip}{2pt}

\title{Model Partition and
Resource Allocation for Split Learning in Vehicular Edge Networks
}
	\author{Lu Yu, Zheng Chang,~\IEEEmembership{Senior Member, IEEE,} Yunjian Jia~\IEEEmembership{Senior Member,~IEEE,}, and Geyong~Min,~\IEEEmembership{Senior Member,~IEEE,}  
   	

        \thanks{Lu Yu and Zheng Chang are with School of Computer Science and Engineering, University of Electronic Science and Technology of China, Chengdu 611731, China. }
        \thanks{Y. Jia is the School of Microelectronics and Communication Engineering, Chongqing University, Chongqing, China } 
        \thanks{G. Min is with Department of Computer Science, University of Exeter, Exeter, EX4 4QF, U.K..}
	}

\markboth{Journal of \LaTeX\ Class Files,~Vol.~14, No.~8, August~2024}%
{Shell \MakeLowercase{\textit{et al.}}: A Sample Article Using IEEEtran.cls for IEEE Journals}
\maketitle

\begin{abstract}
The integration of autonomous driving technologies with vehicular networks presents significant challenges in privacy preservation, communication efficiency, and resource allocation. This paper proposes a novel U-shaped split federated learning (U-SFL) framework to address these challenges on the way of realizing in vehicular edge networks. U-SFL is able to enhance privacy protection by keeping both raw data and labels on the vehicular user (VU) side while enabling parallel processing across multiple vehicles. To optimize communication efficiency, we introduce a semantic-aware auto-encoder (SAE) that significantly reduces the dimensionality of transmitted data while preserving essential semantic information. Furthermore, we develop a deep reinforcement learning (DRL) based algorithm to solve the NP-hard problem of dynamic resource allocation and split point selection. Our comprehensive evaluation demonstrates that U-SFL achieves comparable classification performance to traditional split learning (SL) while substantially reducing data transmission volume and communication latency. The proposed DRL-based optimization algorithm shows good convergence in balancing latency, energy consumption, and learning performance. 
\end{abstract}

\IEEEpeerreviewmaketitle

\section{Introduction}
\IEEEPARstart{A}{utonomous} driving technology stands at the forefront of automotive innovation, promising to revolutionize transportation systems and reshape urban mobility \cite{autonomousdriving_review}. As vehicles evolve towards higher levels of autonomy, the potential for enhanced road safety, improved traffic efficiency, and reduced environmental impact becomes increasingly apparent \cite{ad_benefits}. Central to the realization of comprehensive autonomous driving is the concept of vehicular networks (VNs), which facilitate crucial vehicle-to-vehicle (V2V) and vehicle-to-infrastructure (V2I) communications \cite{v2x_overview}.
The integration of autonomous driving technologies with vehicular networks, while promising, presents significant challenges that must be addressed to ensure robust and efficient operation. These challenges primarily stem from the unique characteristics of the vehicular networks and the stringent requirements of autonomous systems \cite{ad_vn_challenges}. Key issues include the need for real-time processing to ensure safe operation, privacy concerns arising from the continuous exchange of sensitive information \cite{privacy_concerns}, and communication efficiency challenges due to the large volume of data generated by autonomous vehicles. \par 


The rapid advancement of autonomous driving technologies have created an urgent need for sophisticated, distributed learning methods tailored to vehicular environment. This necessity is driven by several key factors: privacy concerns, communication overhead, latency requirements and resource constraints. 
Traditional centralized machine learning approaches require the aggregation of vast amounts of data from vehicles, including sensitive information such as location history, driving patterns, and potentially personal data captured by in-vehicle sensors.  
The high mobility and large scale of vehicular networks result in substantial communication overhead when transmitting raw data to central servers \cite{VN_communication_challenges}. This overhead can lead to increased latency, network congestion, and higher operational costs.
While modern vehicles are increasingly equipped with computational resources, they still face limitations in processing power, energy consumption, and storage capacity compared to centralized data centers \cite{vehicle_resource_constraints}. Efficient utilization of these limited resources is crucial for implementing sophisticated ML models in vehicular settings. \par

To address the challenges in autonomous driving and vehicular networks, several distributed learning approaches have been proposed. However, these methods have limitations when applied to the unique environment of vehicular networks.
Federated learning (FL) has emerged as a promising distributed learning paradigm that allows model training on decentralized data \cite{fl_overview}. While FL addresses some privacy concerns, it faces several limitations in vehicular networks. FL requires multiple rounds of model updates, which can be challenging in the high-mobility environment of vehicular networks \cite{fl_comm_challenge}.
FL assumes clients have sufficient computational resources to train local models, which may not always be the case for all vehicles \cite{fl_resource_limitation}.
Traditional split learning (SL) is a distributed learning approach that divides the neural network model between clients and the server, offering a balance between privacy preservation and computational efficiency \cite{SL- first_proposed}. While SL provides several advantages, it faces significant limitations in the context of vehicular networks. Traditional SL operates sequentially, which can lead to inefficiencies in multi-client scenarios typical in vehicular networks \cite{sl_sequential_limitation}. Although SL keeps raw data on clients, the transmitted activations may still leak sensitive information \cite{sl_privacy_concern}. In traditional SL, labels are typically shared with the server, which can lead to privacy leakage through label inference attacks. This is particularly concerning in vehicular networks where labels might correspond to sensitive driving behaviors or locations \cite{sl_label_attack}. The fixed split point in traditional SL may not be optimal for all vehicles due to varying computational capabilities and network conditions. While SL reduces data transmission compared to centralized approaches, it still requires significant communication between clients and the server, which can be challenging in dynamic vehicular environment
\cite{sl_communication_challenge}.
These limitations underscore the need for a more adaptive and efficient distributed learning approach tailored to the unique challenges of autonomous driving in vehicular networks. Such an approach should address the privacy concerns, communication efficiency issues, and resource constraints while leveraging the distributed nature of vehicular networks to enhance learning performance. \par

In the context of vehicular networks, the size of intermediate features in deep neural networks (DNNs) used for autonomous driving tasks often exceeds that of the original input data, the problem is further exacerbated in multi-vehicle environments. This challenge necessitates the development of effective feature compression techniques to reduce transmission overhead.
In this context, semantic communication emerges as a promising solution. By focusing on transmitting the semantic meaning of data rather than raw bits, semantic communication can significantly reduce the amount of data transmitted while preserving essential information. This approach is particularly relevant in vehicular networks, where the semantic content of data (e.g., road conditions, traffic patterns, or obstacle detection) is often more crucial than the exact pixel values of images or precise numerical readings from sensors. Integrating semantic communication techniques into the SL framework could potentially address both the challenges of feature compression and the preservation of critical information for autonomous driving tasks. 
Meanwhile, a single edge server (ES) typically serves multiple vehicles, with these vehicles communicating through a shared channel. Even with intermediate features compression, the training between vehicles can still lead to significant additional latency. To address this issue, it becomes crucial to optimize resource allocation, including bandwidth and computational resources. However, this optimization problem is typically NP-hard \cite{resource_allocation_np_hard}, making it challenging to find optimal solutions in real-time, especially in the dynamic environment of vehicular networks. \par

In this paper, we propose a novel U-shaped split federated learning (U-SFL) framework, specifically tailored for autonomous driving applications in vehicular networks. This framework builds upon the foundations of traditional SL while introducing key innovations to address its limitations. By integrating advanced techniques such as semantic-aware auto-encoder (SAE) and deep reinforcement learning (DRL) for resource allocation, U-SFL aims to provide a comprehensive solution that balances privacy preservation, communication efficiency, and learning performance in the challenging context of vehicular networks and autonomous driving. 
Our main contributions are as follows:
\begin{itemize}
\item We propose a novel U-SFL framework that enhances privacy protection and enables parallel processing while maintaining comparable classification performance to traditional SL. The U-shaped architecture allows for efficient distribution of the learning process across vehicles and ES while keeping both raw data and labels on the vehicle side, significantly improving privacy compared to traditional SL approaches.
\item We introduce a SAE component into the U-SFL framework to improve communication efficiency. By leveraging the principles of semantic communication, the SAE reduces the dimensionality of transmitted data while preserving important semantic information, thereby minimizing communication overhead in bandwidth-constrained vehicular networks. This integration of semantic communication principles with SL represents a significant advancement in addressing the unique challenges of vehicular edge computing.
\item We develop a sophisticated DRL-based algorithm to solve the NP-hard problem of dynamic resource allocation and split point selection in the context of vehicular networks.    Our approach addresses the complex, high-dimensional state space that includes vehicle locations, network conditions, and computational loads, while managing a hybrid action space that combines discrete decisions (split point selection) with continuous actions (resource allocation). This DRL-based solution optimizes the utilization of shared resources across multiple vehicles in a realistic, time-varying vehicular network setting. 
\par
\end{itemize}  

The rest of this paper is organized as follows: Section II reviews related work. Section III presents the system model, the proposed U-SFL framework and the semantic-aware communication techniques. Section IV details the computation and communication modeling. Section V introduces the DRL-based multi-objective optimization algorithm. Section VI presents and discusses the simulation results. Finally, Section VII concludes the paper and outlines future research directions. \par

\section{Related Work}
SL has emerged as a promising distributed learning paradigm \cite{SL- first_proposed} that addresses privacy concerns while enabling collaborative model training in resource-constrained environments such as vehicular networks \cite{sl_label_attack}. 
There are several variants of SL. Its original form, called vanilla SL, targets privacy-preserving healthcare system \cite{SL-health}. It operates in a sequential manner, training the model for one client at a time. However, the sequential training process of vanilla SL incurs excessive training latency. From the communication perspective, SL is slower than FL because FL is trained in parallel. 
To address these issues, split federated learning (SFL) \cite{SL-SFL1}-\cite{SL-SFL3} and parallel split learning (PSL) \cite{SL-PSL1}, \cite{SL-PSL2} have been devised to parallelize client-side model training, empowering clients to train their sub-models simultaneously.
These approaches aim to leverage the advantages of both paradigms, potentially improving efficiency in multi-client scenarios typical in vehicular networks. 
There are also some studies that proposed to use of a global server to aggregate the multiple client-side models and the server-side models \cite{SL-aggregate1}, \cite{SL-aggregate2}, which is similar to the proposed paradigm in our work.
While these hybrid approaches offer some improvements, they still face significant challenges. Notably, they do not fully address the potential privacy leakage through label sharing, which remains a critical concern in sensitive applications like autonomous driving. Moreover, the communication overhead in these approaches remains substantial, particularly in bandwidth-constrained vehicular environments.

Resource efficiency is a critical concern in the application of split learning to vehicular networks, where computational resources, communication bandwidth, and energy are often constrained \cite{GANPowered}.
First of all, it is of paramount importance to reduce communication overhead for smashed data exchange between vehicles and the edge server. To mitigate this issue, one promising direction is to adopt auto-encoder, which trains an encoder to compress the data and then a decoder to recover the data \cite{SL-auto_encoder1}, \cite{SL-auto_encoder2}. But auto-encoder will bring additional computation and training costs.
In parallel split learning, the training latency is determined by the slowest client, also known as the ``straggler". To mitigate this issue, the channels and server-side computing resources should be judiciously allocated to the stragglers to optimize the training process \cite{SL-straggler}.
Network resource allocation is tightly coupled with model splitting in SL. The split layer significantly impacts training latency, as it leads to varying training workloads between end devices and edge servers and different communication overheads due to the output data sizes across layers.
Along this line, some studies propose a cluster-based SL in which clients concurrently train the model in each cluster based on SFL \cite{SL-cluster1}. Subsequently, the model undergoes training across different groups based on the traditional SL method. This approach stochastically optimizes the cut layer selection, device clustering, and radio spectrum allocation \cite{SL-cluster2}.

While these approaches offer valuable insights, they lack a comprehensive framework that jointly optimizes model partitioning, resource allocation, communication efficiency and privacy preserving in the context of vehicular networks. Our proposed U-SFL framework, integrated with SAE and DRL-based optimization, addresses these limitations by providing a holistic solution that balances privacy preservation, communication efficiency, and resource utilization in vehicular edge networks.

\section{System model}
\subsection{Vehicular Network Architecture}
\begin{figure}[t]
    \centering
    \renewcommand{\figurename}{Fig.}
    \includegraphics[width=7cm]{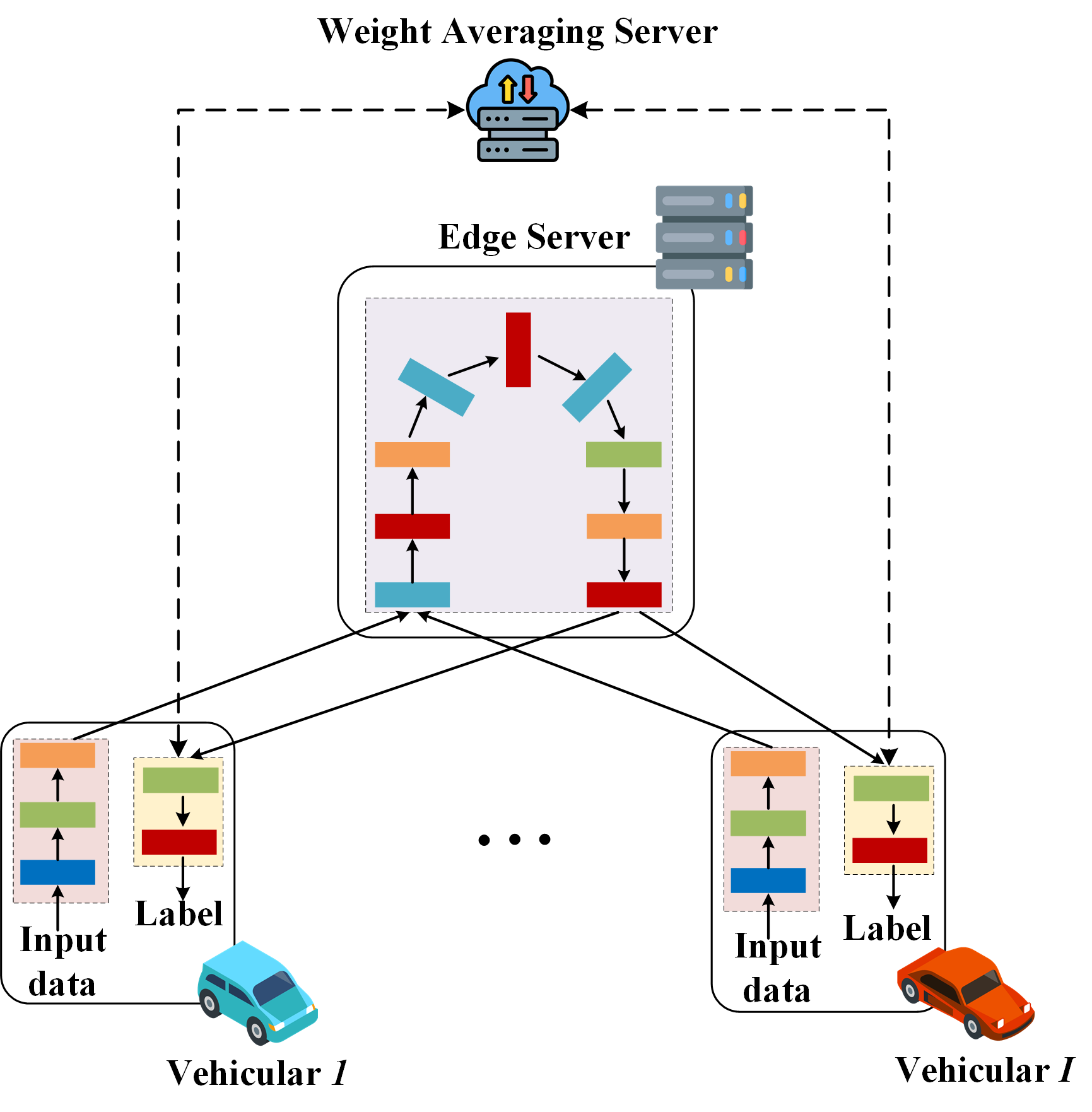}
    \caption{System model.}
    \label{fig:systemmodel}
\end{figure}

We consider a vehicular edge computing network comprising $I$ vehicular users (VUs), a weight averaging server (WAS) and one edge server (ES), as illustrated in Fig. \ref{fig:systemmodel}. 
The set of VUs is denoted as $\mathcal{I}=\{1,2,\ldots,I\}$, where each VU $i\in\mathcal{I}$ possesses local computational capabilities and a dataset $\{(\boldsymbol {x}_{i,1},y_{i,1}),\ldots,(\boldsymbol {x}_{i,D_i},y_{i,D_i})\}$ with the size $D_i$, where $\boldsymbol {x}_{i,1},\ldots,\boldsymbol {x}_{i,D_i}$ are the raw data and $y_{i,1},\ldots,y_{i,D_i}$ refer to the corresponding labels. The aggregate dataset across all VUs is defined as $D=\sum_{i\in\mathcal{I}}D_i$.

Each VU $i \in \mathcal{I}$ is equipped with a semantic encoder, which is responsible for extracting and compressing the semantic information from the raw data before transmission. This semantic encoding process is crucial for reducing the communication overhead while preserving the essential information needed for the learning task. 
The WAS periodically aggregates the model parameters from all VUs, performing a federated averaging operation to enhance model consistency across the network while preserving privacy. The WAS does not have access to raw data and only deals with model parameters, further reinforcing the privacy-preserving nature of our U-SFL framework.
The ES, endowed with superior computational resources, facilitates the distributed learning process. It maintains a portion of the neural network model, coordinates the learning activities across all VUs, and incorporates a semantic decoder to reconstruct the semantic information received from the VUs. This semantic-aware communication paradigm enables efficient data exchange in the bandwidth-constrained vehicular network environment. 
This architecture, integrating U-SFL with semantic-aware communication, enables the implementation of our proposed framework, which we will elaborate on in subsequent sections.  

\subsection{U-shaped Split Federated Learning Model}
Building upon the network architecture described previously, we propose a novel U-SFL model. 
U-SFL leverages the distributed nature of the network to partition the deep learning model across VUs and the ES in U-shaped configuration. This novel approach enables efficient collaborative learning while addressing the unique challenges of vehicular networks, such as privacy preservation and resource constraints.

\subsubsection{Model Structure}
\begin{figure}[t]
    \centering
    \renewcommand{\figurename}{Fig.}
    \includegraphics[width=7cm]{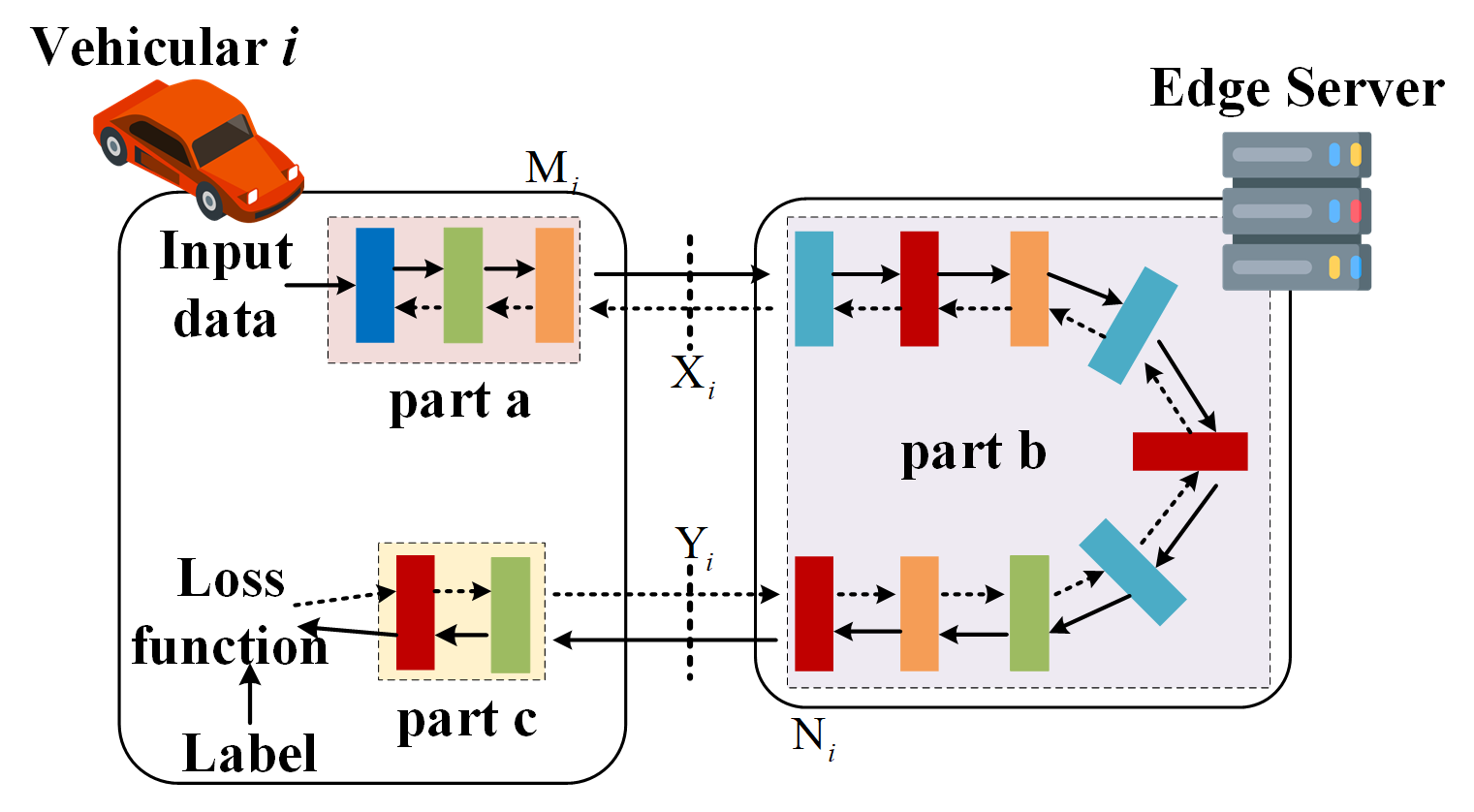}
    \caption{U-SFL scheme without label sharing, the model is split into three parts (part a, part b and part c) for training.}
    \label{fig:U-SFL}
\end{figure}

In the U-SFL scheme, the neural network is divided into three parts, as illustrated in Fig. \ref{fig:U-SFL}: the initial layers (part a) and final layers (part c) are computed on the VUs, while the intermediate layers (part b) are processed on the ES. 
This U-shaped structure allows for privacy-preserving feature extraction and classification on the VUs, with complex intermediate computations offloaded to the ES. 
Formally, we denote the model parameters as $\boldsymbol{\omega } = \{\boldsymbol{\omega }_{a}^i, \boldsymbol{\omega }_{b}, \boldsymbol{\omega }_{c}^i\}$, where $\boldsymbol{\omega }_{a}^i$ and $\boldsymbol{\omega }_{c}^i$ are the parameters for the local parts of VU $i$, and $\boldsymbol{\omega }_{b}$ represents the parameters for the ES.

\subsubsection{Training Process}

The U-SFL training process involves four key steps.
\textbf{\emph{Forward Propagation}}: VUs process local data through part a, send activations $\mathbf{M}_i$ to ES, which processes through part b to produce $\mathbf{N}_i$, then VUs complete forward pass through part c.
\textbf{\emph{Backward Propagation}}: VUs compute loss and initiate backpropagation through part c, ES continues through part b, and VUs complete through part a.
\textbf{\emph{Parameter Update}}: VUs update local parameters ($\boldsymbol{\omega }_{a}$ and $\boldsymbol{\omega }_{c}$); ES updates its parameters ($\boldsymbol{\omega }_{b}$).
\textbf{\emph{Model Aggregation}}: Periodically, the WAS aggregates local models across all VUs.
The overall process is summarized in Algorithm \ref{algorithm1}.

\begin{algorithm}[t]
\caption{U-shaped Split Federated Learning (U-SFL) Scheme}
\label{algorithm1}
\begin{algorithmic}[1]
\Require $I$ VUs, initial model parameters $\boldsymbol{\omega } = \{\boldsymbol{\omega }_{a}^i, \boldsymbol{\omega }_{b}, \boldsymbol{\omega }_{c}^i\}$, learning rate $\eta$, local epochs $E$, global iterations $G$, batch size $B$
\Ensure Trained model parameters $\boldsymbol{\omega }^{G} = \{\boldsymbol{\omega }_{a}^{G,i}, \boldsymbol{\omega }_{b}^{G}, \boldsymbol{\omega }_{c}^{G,i}\}$
    \For{global iteration $g = 1$ to $G$}
        \For{local epoch $e = 1$ to $E$}
            \ForAll{VUs $i \in I$ in parallel}
                \State $\mathbf{M}_i = f_{a}(\xi_i; \boldsymbol{\omega }_{a}^i)$ \Comment{Forward prop part a}
                \State Send $\mathbf{M}_i$ to ES
            \EndFor
            \State ES: $\mathbf{N}_i = f_{b}(\mathbf{M}_i; \boldsymbol{\omega }_{b})$ for each VU $i$
            \ForAll{VUs $i \in I$ in parallel}
                \State $\mathbf{N}_{\text{output}}^i = f_{c}(\mathbf{N}_i; \boldsymbol{\omega }_{c}^i)$ \Comment{Complete forward prop}
                \State Compute loss $L^i$ and backpropagate
                \State Update $\boldsymbol{\omega }_{a}^i$, $\boldsymbol{\omega }_{c}^i$
            \EndFor
            \State ES: Update $\boldsymbol{\omega }_{b}$
        \EndFor
        \If{$g \mod E == 0$}
                \State Aggregate: $\boldsymbol{\omega }_{a} = \frac{1}{I} \sum_{i=1}^{I} \boldsymbol{\omega }_{a}^i$, $\boldsymbol{\omega }_{c} = \frac{1}{I} \sum_{i=1}^{I} \boldsymbol{\omega }_{c}^i$
                \State Distribute aggregated parameters to all VUs
        \EndIf
    \EndFor
\State\Return $\boldsymbol{\omega }^{G}$
\end{algorithmic}
\end{algorithm}

\subsubsection{Privacy and Efficiency Considerations}
The U-SFL model addresses privacy concerns by keeping raw data and labels on local devices. Only intermediate representations are transmitted, which are more difficult to reverse the original data. Additionally, the U-shaped structure allows for efficient utilization of both VU and ES resources, potentially reducing energy consumption and latency.
By carefully selecting the split points (denoted as $X_i$ and $Y_i$) in the neural network and allocating communication resources, we aim to optimize the trade-off between computational load distribution and communication efficiency. This optimization problem will be formulated in detail in the subsequent section.

\subsection{Semantic-aware Communication for U-SFL}
\begin{figure}[t]
    \centering
    \renewcommand{\figurename}{Fig.}
    \includegraphics[width=7cm]{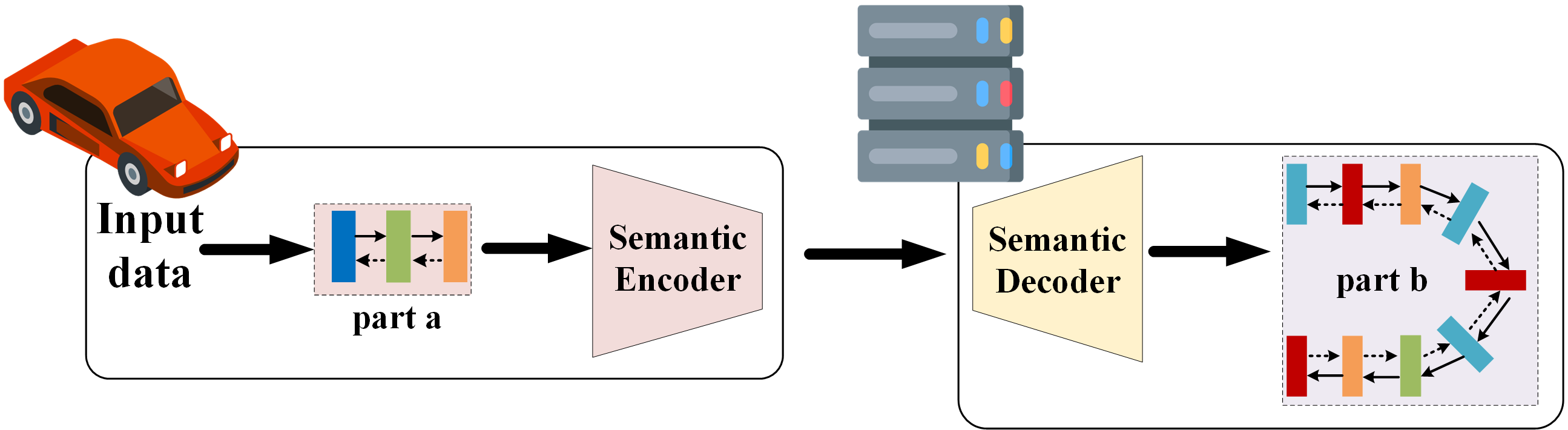}
    \caption{Semantic-aware communication framework for U-SFL in vehicular edge networks.}
    \label{fig:SemCom_NoChannel}
\end{figure}

To enhance the efficiency of our U-SFL scheme in vehicular networks, we introduce semantic-aware communication techniques. This approach aims to reduce the amount of data transmitted between VUs and the ES without significantly compromising the learning performance. Fig. \ref{fig:SemCom_NoChannel} illustrates the overall structure of our semantic-aware communication framework integrated into the U-SFL scheme. As shown in Fig. \ref{fig:SemCom_NoChannel}, the framework consists of three main components: (1) the input data processing at the vehicular user side (part a), (2) the semantic encoder and decoder for efficient communication, and (3) the server-side processing (part b) followed by the final processing at the vehicular user side (part c). 

\subsubsection{Semantic Information Completeness}
We define semantic information completeness as the degree to which the intermediate representations capture the essential features for the classification task. For a given layer $l$, we quantify this using a semantic completeness score $S(l)$:
\begin{small}
\begin{equation}
\label{eq_SemInfoCompleteness}
S(l) = f(I(\mathcal{Q}_l; \mathcal{C})),
\end{equation}
\end{small}where $I(\mathcal{Q}_l; \mathcal{C})$ represents the mutual information between the layer output $\mathcal{Q}_l$ and the final classification $\mathcal{C}$, and $f(\cdot)$ is a monotonically increasing function.

\subsubsection{Semantic-Sensitive Split Point Selection}
In U-SFL, the choice of split points $X$ and $Y$ significantly impacts both communication efficiency and semantic information preservation. We focus primarily on semantic transmission at the first split point $X$, i.e., between part a and part b. We propose a semantic-sensitive split point selection strategy:
\begin{small}
\begin{equation}
\label{eq_SemPointSelect}
X^* = \arg\max_{l} {S(l) | l \in [1, L/2]},
\end{equation}
\end{small}where $L$ is the total number of layers in the neural network. This approach tightly integrates semantic communication with split point selection, focusing on splitting at layers where semantic information is relatively complete, thereby improving communication efficiency without significantly degrading performance.

\subsubsection{Semantic-aware Feature Compression}
In our U-SFL framework, we implement semantic-aware feature compression using a semantic encoder at the transmitter side (VUs) and a semantic decoder at the receiver side (ES). This approach allows us to extract and transmit only the most relevant semantic information, thereby reducing the communication overhead.

\begin{itemize}
\item \textbf{Semantic Encoder}:
The semantic encoder is responsible for extracting the semantic information from the source data. Let the source information be denoted by $K$. The semantic encoder $T_\beta(\cdot)$, parameterized by $\beta$, processes the source information to produce a semantic representation $A$. This can be expressed as:
\begin{small}
\begin{equation}
\label{eq_SemEncoder}
A = T_\beta(K),
\end{equation}
\end{small}where $A$ represents the compressed semantic features to be transmitted over the network.

\item \textbf{Semantic Decoder}:
At the receiver side, the semantic decoder $R_\varphi(\cdot)$, parameterized by $\varphi$, processes the received semantic representation to reconstruct the original information. The restored information $\hat{K}$ is obtained as:
\begin{small}
\begin{equation}
\label{eq_SemDecoder}
\hat{K} = R_\varphi(A).
\end{equation}
\end{small}

\item \textbf{Training Objective}:
The semantic encoder and decoder are trained to minimize the reconstruction error while maximizing the compression ratio. We use the Mean Squared Error (MSE) as the loss function:
\begin{small}
\begin{equation}
\label{eq_SemLoss}
L_{MSE}(\beta, \varphi) = \frac{1}{N}\sum_{i=1}^{N} (K_i - \hat{K}_i)^2,
\end{equation}
\end{small}where $N$ is the number of samples, $K_i$ is the original information, and $\hat{K}_i$ is the reconstructed information.

\end{itemize}

By integrating this semantic-aware feature compression into our U-SFL framework, we can significantly reduce the amount of data transmitted between VUs and the ES, while preserving the essential semantic information required for the learning task.



\section{Computation And Communication Modeling With Problem Formulation}
\subsection{Computation Model}
In our U-SFL framework with semantic-aware communication, the computation model needs to account for the processing at both the VUs and the ES, including the semantic encoding and decoding processes. Let $X_i$ and $Y_i$ be the split decision of the VU $i$, the DNN can be split into three parts, namely the input layer to the $X_i$-th layer, the ($X_i$ +1)-th layer to the $Y_i$-th layer, and the ($Y_i$ + 1)-th layer to the output layer, which are denoted as part a, part b and part c, respectively.

\subsubsection{VU Computation}
The computation time for VU $i$ consists of three components, they are \textbf{\emph{part a computation}}, \textbf{\emph{semantic encoding}} and \textbf{\emph{part c computation}}.

\textbf{\emph{Part a computation}}:
\begin{small}
\begin{equation}
\label{eq_cmpa}
T_{v,i}^{cmpa}=\frac{\sum_{l=1}^{X_i}b_i(F_l^F+F_l^B)}{f_{v,i}^{cmp}n_i},
\end{equation}
\end{small}where $b_i$ is the batch size of VU $i$, $F^F_l$ and $F^B_l$ are the number of floating point operations
(FLOPs) required by the $l$-th layer in forward and backward propagation, respectively, $f_{v,i}^{cmp}$ is the CPU clock frequency, and $n_i$ is the number of CPU FLOPs per cycle. Thus, vehicular $i$’s computing capability is $f_{v,i}^{flops}=f_{v,i}^{cmp}n_i$. 

\textbf{\emph{Semantic encoding}}:
\begin{small}
\begin{equation}
\label{eq_SemEnc}
T_{v,i}^{SemEnc} = \frac{b_i F^{SemEnc}}{f_{v,i}^{cmp} n_i},
\end{equation}
\end{small}where $F^{SemEnc}$ is the number of FLOPs required for semantic encoding.

\textbf{\emph{Part c computation}}:
\begin{small}
\begin{equation}
\label{eq_cmpc}
T_{v,i}^{cmpc}=\frac{\sum_{l=Y_i+1}^{L}b_i(F_l^F+F_l^B)}{f_{v,i}^{cmp}n_i}.
\end{equation}
\end{small}where $L$ is the total number of layers of the neural network.

\subsubsection{ES Computation}
The computation time at the ES for VU $i$ includes \textbf{\emph{semantic decoding}} and \textbf{\emph{part b computation}}.
 
\textbf{\emph{Semantic decoding}}:
\begin{small}
\begin{equation}
\label{eq_SemDec}
T_{e,i}^{SemDec} = \frac{b_i F^{SemDec}}{f_{e,i}^{cmp}n_e},
\end{equation}
\end{small}where $F^{SemDec}$ is the number of FLOPs for semantic decoding, $f_{e,i}^{cmp}$ is the CPU clock frequency allocated to VU $i$, and $n_e$ is the ES's CPU FLOPs per cycle.

\textbf{\emph{Part b computation}}:
\begin{small}
\begin{equation}
\label{eq_cmpb}
T_{e,i}^{cmpb}=\frac{\sum_{l=X_i+1}^{Y_i}b_i(F_l^F+F_l^B)}{f_{e,i}^{cmp}n_e}.
\end{equation}
\end{small} 

\subsubsection{Energy Consumption}
According to \cite{sl_communication_challenge}, the CPU's power consumption of vehicular device $i$ is given as $P_i=\psi_i(f_{v,i}^{cmp})^3$ where $\psi_i$ is the coefficient [in $Watt/(Cycle/s)^3]$ according to the chip architecture. The energy consumption for computation at VU $i$ regarding the computation at each communication round can be given as
\begin{small}
\begin{equation}
\label{eq_cmpVUEnergy}
E_i^{cmp}=P_i^{cmp}(T_{v,i}^{cmp})=\psi_i(f_{v,i}^{cmp})^3(T_{v,i}^{cmpa}+T_{v,i}^{SemEnc}+T_{v,i}^{cmpc}).
\end{equation}
\end{small}

\subsection{Communication Model}
In our semantic-aware U-SFL framework for vehicular networks, we consider a dynamic communication model that accounts for the mobility of VUs. Our model incorporates the mobility of VUs through a time-varying distance function, which directly impacts the channel conditions and data rates. By using average data rates over a VU's stay within the ES's coverage, we capture the essence of mobility while maintaining tractability in our optimization problem. This approach allows us to consider the dynamic nature of vehicular networks while focusing on the benefits of our semantic-aware U-SFL framework.

\subsubsection{Channel Model}
For VU $i$, we define the time-varying distance to the ES as:
\begin{small}
\begin{equation}
\label{eq_distance}
d_i(t)=\begin{cases}\sqrt{d_h^2+(d_c/2-l_i^0-V_it)^2},&\text{if }l_i^0\leq d_c/2,\\\sqrt{d_h^2+(l_i^0-d_c/2+V_it)^2},&\text{if }l_i^0>d_c/2,\end{cases}
\end{equation}
\end{small}where $d_h$ is the height of the ES, $d_c$ is the coverage diameter of the ES, $l_i^0$ is the initial location of VU $i$. We assume that VU $i$ travels across the edge server at a constant speed $V_i$.

The uplink and downlink data rates at time $t$ are given by:
\begin{small}
\begin{equation}
\label{eq_uplinkRate}
r_i^U(t) = B_i \log_2 \left(1 + \frac{P_i h_0 d_i(t)^{-\alpha}}{N_0}\right),
\end{equation}
\end{small}

\begin{small}
\begin{equation}
\label{eq_DownlinkRate}
r_i^D(t) = B_D \log_2 \left(1 + \frac{P_E h_0 d_i(t)^{-\alpha}}{N_0}\right),
\end{equation}
\end{small}where $B_i$ is the allocated uplink bandwidth, $B_D$ is the downlink bandwidth, $P_i$ and $P_E$ are the transmit powers of VU $i$ and the ES respectively, $h_0$ is the channel gain at unit distance, $\alpha$ is the path loss exponent, and $N_0$ is the noise power spectral density.

\subsubsection{Average Data Rate}
To account for the varying channel conditions during the stay of VU $i$ in the ES's coverage area, we consider the average data rates:
\begin{small}
\begin{equation}
\label{eq_avgUplinkRate}
\bar r_i^U = \frac{1}{t_{i,stay}} \int_0^{t_{i,stay}} r_i^U(t) dt,
\end{equation}
\end{small}

\begin{small}
\begin{equation}
\label{eq_avgDownlinkRate}
\bar r_i^D=\frac{1}{t_{i,stay}} \int_{0}^{t_{i,stay}} r_i^D(t)dt,
\end{equation}
\end{small}where $t_{i,stay}$ is the duration of VU $i$'s stay within the ES's coverage area.

\subsubsection{Transmission Latency}
Uplink transmission of semantically encoded data:
\begin{small}
\begin{equation}
\label{eq_comFa}
T_{v,i}^{comF_a} = \frac{b_i O_{X_i}^{sem}}{\bar{r}_i^U},
\end{equation}
\end{small}where $O^{sem}_{X_i}$ is the size of the semantically encoded output from part a.

Downlink transmission of part b output:
\begin{small}
\begin{equation}
\label{eq_comFb}
T_{e,i}^{comF_b} = \frac{b_i O^F_{Y_i}}{\bar{r}_i^D},
\end{equation}
\end{small}where $O^F_{Y_i}$ is the size of the output from part b.

Uplink transmission of gradients from part c:
\begin{small}
\begin{equation}
\label{eq_comBc}
T_{v,i}^{comB_c} = \frac{b_i O^B_{Y_i+1}}{\bar{r}_i^U},
\end{equation}
\end{small}where $O^B_{Y_i+1}$ is the size of the gradients from part c.

Downlink transmission of gradients for part a:
\begin{small}
\begin{equation}
\label{eq_comBb}
T_{e,i}^{comB_b} = \frac{b_i O^B_{X_i+1}}{\bar{r}_i^D},
\end{equation}
\end{small}where $O^B_{X_i+1}$ is the size of the gradients for part a.

\subsubsection{Communication Energy Consumption}
The energy consumption for communication at VU $i$ is:
\begin{small}
\begin{equation}
\label{eq_comEnergy}
E_i^{com} = P_i(T_{v,i}^{comF_a} + T_{v,i}^{comB_c}) + P_{i, r}(T_{e,i}^{comF_b} + T_{e,i}^{comB_b}),
\end{equation}
\end{small}where $P_{i, r}$ is the power consumption of VU $i$ when it receives the data.

This model effectively combines the mobility aspects of vehicular networks with the semantic communication framework. It accounts for the time-varying nature of the channel while still focusing on the semantic aspects of the communication.

\subsection{Problem Formulation}
Our objective is to jointly optimize the model partition, resource allocation to minimize the overall system cost in each communication round, which includes both latency and energy consumption. The problem can be formulated as follows:
\begin{small}
\begin{alignat}{1}
\label{eq_optimization}
&(P1) 
\min_{\mathbb{B},\mathbb{F},\mathbb{X},\mathbb{Y}}\quad\sum_{i=1}^IE_i^{total}+\rho\max\{T_i^{total}\} \notag\\
\mathrm{s.t.}
&\text{C1:}\,\,\,\, \sum_{i=1}^{I} B_i \leq B^{total}, \notag \\
&\text{C2:}\,\,\,\, \sum_{i=1}^{I} f_{e,i}^{cmp} \leq F_e^{total}, \notag \\
&\text{C3:}\,\,\,\,  1 \leq X_i < Y_i \leq L,\quad \forall i \in \mathcal{I}, X_i, Y_i\in\mathbb{Z}, \notag \\
&\text{C4:}\,\,\,\, T_i^{total} \leq t_{i,stay}, \notag \\
&\text{C5:}\,\,\,\, E_i^{total} \leq E_i^{max},
\end{alignat}
\end{small}where:
$\mathbb{B} = {B_1, ..., B_I}$ is the set of bandwidth allocation decisions.
$\mathbb{F} = {f_{e,1}^{cmp}, ..., f_{e,I}^{cmp}}$ is the set of computational frequency allocation decisions at the ES.
$\mathbb{X} = {X_1, ..., X_I}$ and $\mathbb{Y} = {Y_1, ..., Y_I}$ are the sets of partition point decisions.
$E_i^{total} = E_i^{cmp} + E_i^{com}$ is the total energy consumption for VU $i$.
$T_i^{total} = T_i^{cmp} + T_i^{com}$ is the total latency for VU $i$.
$\rho$ is a weighting factor balancing the trade-off between energy consumption and latency.

The constraints are interpreted as follows:
C1: The total allocated bandwidth cannot exceed the system bandwidth $B^{total}$.
C2: The total computational resources allocated at the ES cannot exceed its capacity $F_e^{total}$.
C3: The partition points must be in a valid range and order.
C4: The total processing time for each VU must not exceed its stay time in the ES's coverage area.
C5: The total energy consumption for each VU must not exceed its maximum energy budget $E_i^{max}$.
The total computation time $T_i^{cmp}$ and communication time $T_i^{com}$ for VU $i$ are given by:
\begin{small}
\begin{equation}
\label{eq_totalCmp}
T_i^{cmp} = T_{v,i}^{cmpa} + T_{v,i}^{SemEnc} + T_{e,i}^{SemDec} + T_{e,i}^{cmpb} + T_{v,i}^{cmpc},
\end{equation}
\end{small}

\begin{small}
\begin{equation}
\label{eq_totalCom}
T_i^{com} = T_{v,i}^{comF_a} + T_{e,i}^{comF_b} + T_{v,i}^{comB_c} + T_{e,i}^{comB_b}.
\end{equation}
\end{small}

This formulation encapsulates the joint optimization of model partition and resource allocation in the U-SFL framework for vehicular networks. It considers both the computation and communication aspects, as well as the constraints imposed by the vehicular environment.

The problem (P1) is a mixed-integer nonlinear programming (MINLP) problem, which is generally NP-hard \cite{DRL}. In the next section, we will propose an efficient algorithm to solve this problem, taking into account the unique characteristics of our semantic-aware U-SFL framework in vehicular networks.

\section{Drl Based Multi-objective Optimization Algorithm}
In this section, we reformulate the problem as a Markov Decision Process (MDP) to facilitate the optimization process. And we propose a DRL algorithm to solve the multi-agent resource allocation optimization problem. 

\subsection{MDP Reformulation}
An MDP is defined as a tuple $(\mathcal{S};\mathcal{A};P;r)$, where $\mathcal{S}$ is a set of states, $\mathcal{A}$ is a set of actions, $P:\mathcal{S}\times\mathcal{A}\times\mathcal{S}\to\mathbf{R}$ is a probability distribution that depicts the system dynamics, and $r:\mathcal{S}\times\mathcal{A}\times\mathcal{S}\to\mathbf{R}$ is the reward.

\subsubsection{State Space}
The state space $\mathcal{S}$ encompasses all relevant information describing the current system status. In our U-SFL framework, the state $s_t$ at time step $t$ is defined as:
\begin{small}
\begin{equation}
\label{eq_stateSpace}
s_t=\{\mathbf{k}_t,\mathbf{l}_t,\mathbf{e}_t,\mathbf{d}_t\},
\end{equation}
\end{small}where $\mathbf{k}_t = \{k_{t,1}, ..., k_{t,I}\}$ is the number of remaining tasks for each VU. $\mathbf{l}_t = \{l_{t,1}, ..., l_{t,I}\}$ is the remaining execution time for each VU. $\mathbf{e}_t = \{e_{t,1}, ..., e_{t,I}\}$ is the remaining energy for each VU. $\mathbf{d}_t = \{d_{t,1}, ..., d_{t,I}\}$ is the distance from each VU to the ES.
This state representation captures the workload, computational progress, energy constraints, and spatial distribution of VUs, enabling the DRL agent to make informed decisions.

\subsubsection{Action Space}
The action space $\mathcal{A}$ comprises all possible decisions the agent can make. In our problem, the action $a_t$ at time step $t$ is defined as:
\begin{small}
\begin{equation}
\label{eq_actionSpace}
a_t=\{\mathbf{b}_t,\mathbf{f}_t,\mathbf{x}_t,\mathbf{y}_t\},
\end{equation}
\end{small}where $\mathbf{b}_t = \{b_{t,1}, ..., b_{t,I}\}$ is the uplink bandwidth allocated to each VU, $\mathbf{f}_t = \{f_{t,1}, ..., f_{t,I}\}$ is the computational frequency allocated to each VU, $\mathbf{x}_t = \{x_{t,1}, ..., x_{t,I}\}$ is the first partition point for each VU, $\mathbf{y}_t = \{y_{t,1}, ..., y_{t,I}\}$ is the second partition point for each VU.
This action space allows the agent to jointly optimize resource allocation and model partitioning decisions for all VUs simultaneously.

\subsubsection{State Transition}
The state transition probability $P(s_{t+1}|s_t, a_t)$ describes the likelihood of transitioning from the current state $s_t$ to the next state $s_{t+1}$, given action $a_t$. 
In our formulation, we have the following considerations.
For the metric $\mathbf{k}_t$, the higher the number of $\mathbf{k}_t$, the more data the VU needs to process, and thus more computational resources ($\mathbf{f}_t$) and communication resources ($\mathbf{b}_t$) may be needed to accelerate task completion. 
For the $\mathbf{l}_t$ metric, the longer the remaining execution time $\mathbf{l}_t$, the heavier the computational task of the VU, and the need to optimize resource allocation ($\mathbf{b}_t$, $\mathbf{f}_t$) to reduce its execution time. 
For the metric $\mathbf{e}_t$, VUs with less remaining energy need to conserve energy in order to extend their working time, and may need to reduce their energy-intensive operations. 
For the metric $\mathbf{d}_t$, VUs that are farther away have higher data transmission latency and energy consumption, and need to optimize the data transmission scheme.

\subsubsection{Reward Function}
The reward function $r_t$ quantifies the immediate return of taking action $a_t$ in state $s_t$. We define the reward as the negative weighted sum of latency and energy consumption:
\begin{small}
\begin{equation}
\label{eq_reward}
r_t = -\sum_{i=1}^I (E_i(t) + \rho T_i(t)),
\end{equation}
\end{small}where $E_i(t)$ and $T_i(t)$ are the energy consumption and latency of VU $i$ at time step $t$, respectively, and $\rho$ is a balancing factor. This reward function encourages the agent to minimize both energy consumption and latency across all VUs.

By reformulating our original optimization problem as an MDP, we transform it into a sequential decision-making problem amenable to DRL techniques. This formulation allows us to leverage the power of DRL to find optimal resource allocation and model partitioning strategies in the dynamic and complex vehicular edge computing environment of our U-SFL framework.

\subsection{DRL Based Multi-objective Optimization Algorithm}
To solve the formulated MDP, we propose a DRL based multi-objective optimization algorithm, specifically tailored to address the complexities of our U-SFL framework in vehicular edge computing networks. This algorithm leverages DRL techniques to handle multi-agent scenarios with hybrid action spaces, making it particularly well-suited for our problem.

\subsubsection{Multi-Agent Consideration}
In our U-SFL scenario, each VU is modeled as an agent, with the ES serving as a central coordinator. Our algorithm trains multiple agents simultaneously, each learning to make decisions for its corresponding VU while considering the global state and the actions of other agents.

\subsubsection{Hybrid Action Space}
The action space in our problem is hybrid, consisting of both discrete (partition points) and continuous (bandwidth and computational frequency allocation) components. Our algorithm is designed to handle such hybrid action spaces effectively, using separate output branches for discrete and continuous actions in its actor network architecture.



\subsection{Actor-Critic Architecture Design}
Our DRL based multi-objective optimization algorithm employs an actor-critic architecture, which is well-suited for handling the complexities of the multi-agent vehicular edge computing networks. This section details the design of both the actor and critic networks.

\subsubsection{Overall Structure}
The algorithm consists of multiple actor networks (one for each VU) and a single, centralized critic network. This design allows for decentralized execution with centralized training, enabling efficient learning in the multi-agent setting. We denote parameters of the critic network and the actor networks as $\phi$ and $\theta_i$, respectively, where $\theta_i$ denotes the parameters of the corresponding actor network of VU $i$.

\subsubsection{Actor Network}
Each actor network is responsible for generating actions for its corresponding VU. The actor network's structure is as follows:
\begin{itemize}
\item \textbf{Input Layer}: 
Accepts the current state $s_t$ as input.
\item \textbf{Shared Layers}: 
Two fully connected layers with 256 and 128 neurons, respectively. LeakyReLU activation functions and two residual blocks for improved gradient flow to extract features from the input state.
\item \textbf{Attention Mechanism}:
Applied to the output of shared layers to emphasize important features.
\item \textbf{Output Branches}: Two discrete branches for partition points $(x_t, y_t)$. Two continuous branches for bandwidth $b_t$ and computational frequency $f_t$.
\end{itemize}


\subsubsection{Critic Network}
The centralized critic network estimates the value function for the joint state of all VUs. Its structure is as follows:
\begin{itemize}
\item \textbf{Input Layer}: 
Accepts the global state (concatenated states of all VUs) as input.
\item \textbf{Hidden Layers}: 
Two fully connected layers with 256 and 128 neurons, respectively. LeakyReLU activation functions, and two residual blocks.
\item \textbf{Output Layer}: 
A single neuron outputting the estimated state value $V(s_t)$.
\end{itemize}


\subsubsection{Handling Hybrid Action Space}
To effectively handle the hybrid action space, we employ separate output branches in the actor network for discrete and continuous actions:
\begin{itemize}

\item 
For discrete actions (partition points), we use categorical distributions:
\begin{small}
\begin{alignat}{1}
\label{eq_discreteAction}
\pi_{\theta_{i}}^{d}(a_{t,i}^{d}|s_{t})&=\prod_{m=1}^{M}p_{m}(s_{t})I_{\{a_{t,i}^{d}=m\}}, \notag \\&\forall i\in\{1,2,\ldots,I\},\sum_{m=1}^{M}p_{m}(s_{t})=1, 
\end{alignat}
\end{small}where the superscript $d$ denotes the discrete part of the action and $M$ is the number of possible actions.

\item 
For continuous actions (bandwidth and frequency allocation), we use Gaussian distributions:
\begin{small}
\begin{alignat}{1}
\label{eq_continuousAction}
\pi_{\theta_i}^c(a_{t,i}^c|s_t)\sim\mathcal{N}(\mu(s_t),\sigma^2(s_t)),
\end{alignat}
\end{small}where the superscript $c$ denotes the continuous part of the action and ${\mathcal{N}}(\cdot)$ is the probability density function of the Gaussian distribution. In practice, the action can be sampled from the above distributions.
\end{itemize}

This design allows our algorithm to learn optimal policies for both the discrete partitioning decisions and the continuous resource allocation decisions simultaneously. By incorporating attention mechanisms and residual blocks, our network architecture is capable of capturing salient features in the state space more effectively, thereby making more informed decisions in the complex vehicular edge computing networks.

\subsection{Optimization Objectives}
We present the optimization objectives for the critic and actor networks. The critic network aims to fit an unknown state-value function, while the actor networks strive to provide policies that maximize the fitted state value. These optimization objectives guide the critic and actor networks to achieve the following goals: minimizing latency and energy consumption while maximizing the efficiency of the U-SFL framework in the vehicular edge computing networks. Specifically, the objectives should direct the networks to learn effective model partitioning and resource allocation strategies to optimize performance in the dynamic vehicular networks. In the following subsections, we will elaborate on the specific optimization objectives and their mathematical formulations for both the critic and actor networks.

\subsubsection{Critic Objective}
The critic network aims to accurately estimate the state-value function. We define the loss function for the critic network as the mean squared error between the estimated state value and the actual returns:
\begin{small}
\begin{equation}
\label{eq_criticLoss}
\mathcal{L}_c(\phi) = \mathbb{E}_t[||V_\phi^\pi(s_t) - V^{\prime\pi}(s_t)||_2],
\end{equation}
\end{small}where $V_\phi^\pi(s_t)$ is the state value estimated by the critic network with parameters $\phi$ under policy $\pi$, and the expectation $\mathbb{E}_t$ is taken over multiple time steps or samples, allowing for a more stable and representative loss estimation across various states encountered under the current policy.

$V^{\prime\pi}(s_t)$ is the real cumulative reward at state $s_t$ under policy $\pi$, computed as:
\begin{small}
\begin{equation}
\label{eq_realCumulativeReward}
V^{\prime\pi}(s_t) = \sum_{t'=t}^{T(\pi)}\gamma^{t'-t}r_{t'},
\end{equation}
\end{small}where $T(\pi)$ denotes the terminal time step of the episode under policy $\pi$, $\gamma \in [0,1]$ is the discount factor, and $r_{t'}$ is the immediate reward at time step $t'$.

\subsubsection{Actor Objective}
For the actor networks, we formulate the actor objective based on advanced policy optimization techniques, aiming to maximize the following:
\begin{small}
\begin{equation}
\label{eq_newOld}
\max_\theta \mathbb{E}_t\left[\frac{\pi_\theta(a_t|s_t)}{\pi_{\theta_{old}}(a_t|s_t)}\hat{A}_t\right],
\end{equation}
\end{small}this objective function leverages the concept of importance sampling, where $\pi_\theta(a_t|s_t)$ represents the current policy, while $\pi_{\theta_{old}}(a_t|s_t)$ is the old policy. The advantage function $\hat{A}_t$ is the advantage function which measures how much a specific action $a_t$ is better than the average actions at state $s_t$. To compute the advantage function, we employ generalized advantage estimation (GAE) \cite{GAE}, which provides a balance between bias and variance in the advantage estimates. The GAE is formulated as follows:
\begin{small}
\begin{equation}
\label{eq_gae}
\hat{A}_t=\sum\limits_{t'=t}^{T(\pi)}(\gamma\lambda)^{t'-t}\Big(r_t+\gamma V_\phi^\pi(s_{t+1})-V'^\pi(s_t)\Big),
\end{equation}
\end{small}where $\lambda \in [0,1]$ is a hyperparameter controlling the bias-variance trade-off. It is important to note that if $t+1 > T(\pi)$, we set $V_{\phi}^{\pi}(s_{t+1}) = 0$.


To enhance the stability of our policy updates in the dynamic vehicular networks, we implement the adaptive clipping mechanism, denoting $(\frac{\pi_\theta(a_t|s_t)}{\pi_{\theta_{old}}(a_t|s_t)})$ as $\mathcal{V}_t(\theta)$:
\begin{small}
\begin{equation}
\label{eq_actorLoss}
\mathcal{L}^{CLIP}(\theta) = \mathbb{E}_t\left[\min( \mathcal{V}_t(\theta) \hat{A}_t, 
\text{clip}( \mathcal{V}_t(\theta) , 1-\epsilon, 1+\epsilon)  \hat{A}_t)\right],
\end{equation}
\end{small}where $\epsilon$ is a hyperparameter that controls how $\mathcal{V}_t(\theta)$ can move away from 1.

To encourage exploration and prevent premature convergence to suboptimal policies, we add an entropy bonus to the actor objective:
\begin{small}
\begin{equation}
\label{eq_finalActorLoss}
\mathcal{L}_a(\theta) =\sum_{i=1}^I \{\mathcal{L}^{CLIP}(\theta_i) + \zeta\mathbb{E}_t\left[\mathcal{H}(\pi_{\theta_i})\right]\},
\end{equation}
\end{small}where $\mathcal{H}(\pi_{\theta_i})$ is an entropy bonus that encourages exploration and $\zeta$ is a balancing hyperparameter controlling the strength of the entropy regularization.

By optimizing these objectives, our algorithm learns to make decisions that effectively balance the trade-offs between model partitioning, resource allocation, latency, and energy consumption in the U-SFL framework.

\section{Simulation Results}
\subsection{U-SFL Convergence Performance}
\subsubsection{Dataset and Preprocessing}
To evaluate the effect of our proposed U-SFL method and the impact of the SAE, we conducted a comprehensive set of experiments using the Caltech-101 and CIFAR-10 dataset. 
The primary objective was to assess the classification performance under various configurations of SL and semantic encoding.
The Caltech-101 dataset, comprising 101 object categories with 40 to 800 images per category, the images are of varying sizes and resolutions. 
The CIFAR-10 dataset, consisting of 60,000 32x32 color images in 10 classes, with 6,000 images per class. There are 50,000 training images and 10,000 test images. 
We employ different learning rates for each dataset: 0.0001 for Caltech-101, 0.00001 for CIFAR-10, optimizer is Adam, batch size is 64, number of epochs is 100.

\begin{figure*}[ht!]
    \centering
    \renewcommand{\figurename}{Fig.}
    \includegraphics[width=18cm]{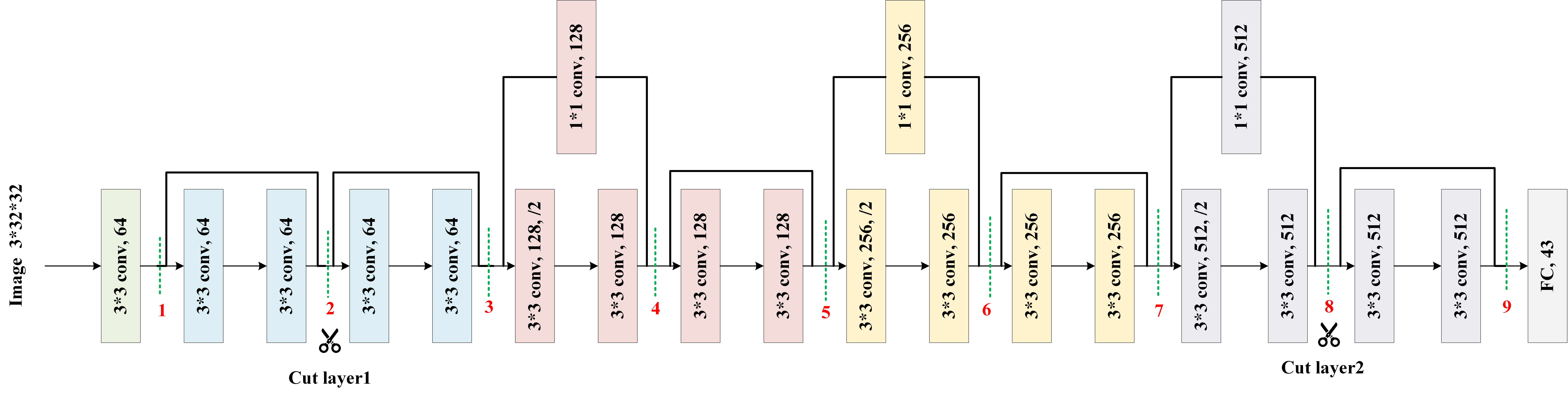}
    \caption{Model structure and partition points for the deep
learning model.}
    \label{fig:ResNet-18}
\end{figure*}

\begin{figure*}[t]
    \centering
    \renewcommand{\figurename}{Fig.}
    \begin{tabular}{ccc}
        \subfigure[Caltech-101 (Training)]{
            \includegraphics[width=0.3\textwidth]{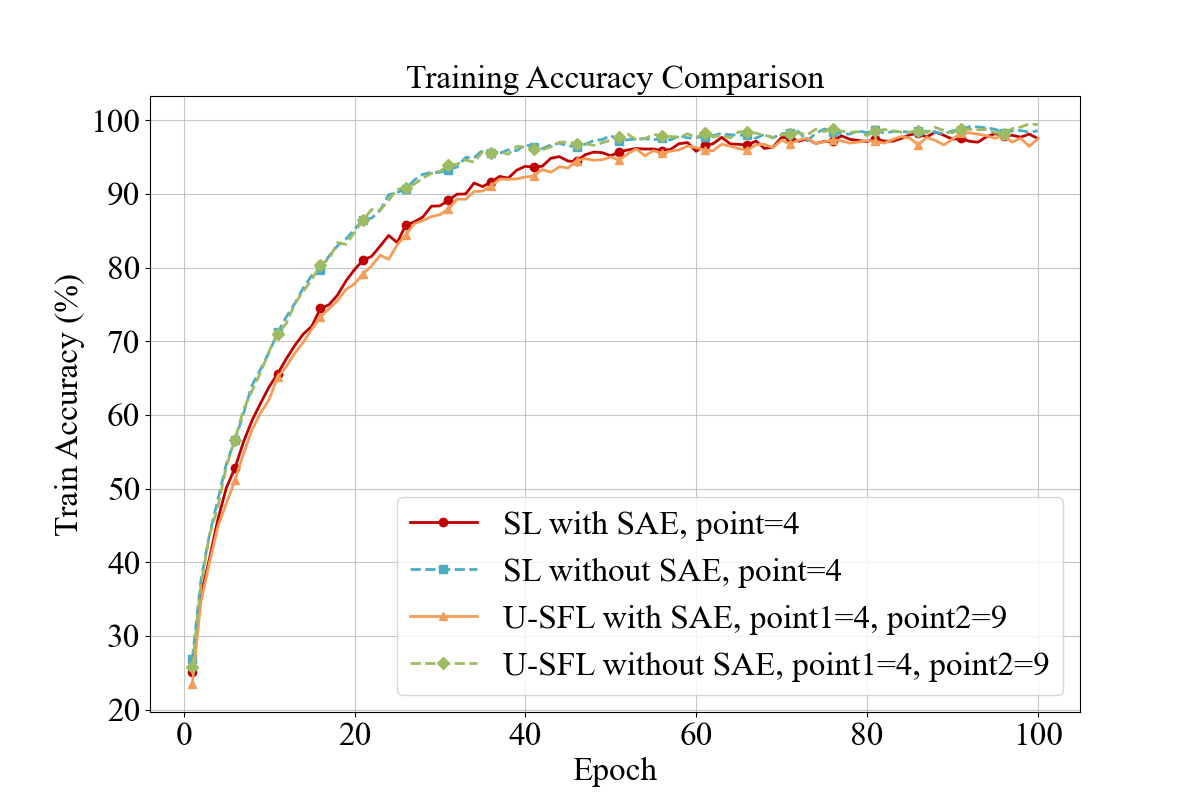}
            \label{fig:Caltech-101_withSAE_withoutSAE_training}
        } &
        \subfigure[Caltech-101 (Testing)]{
            \includegraphics[width=0.3\textwidth]{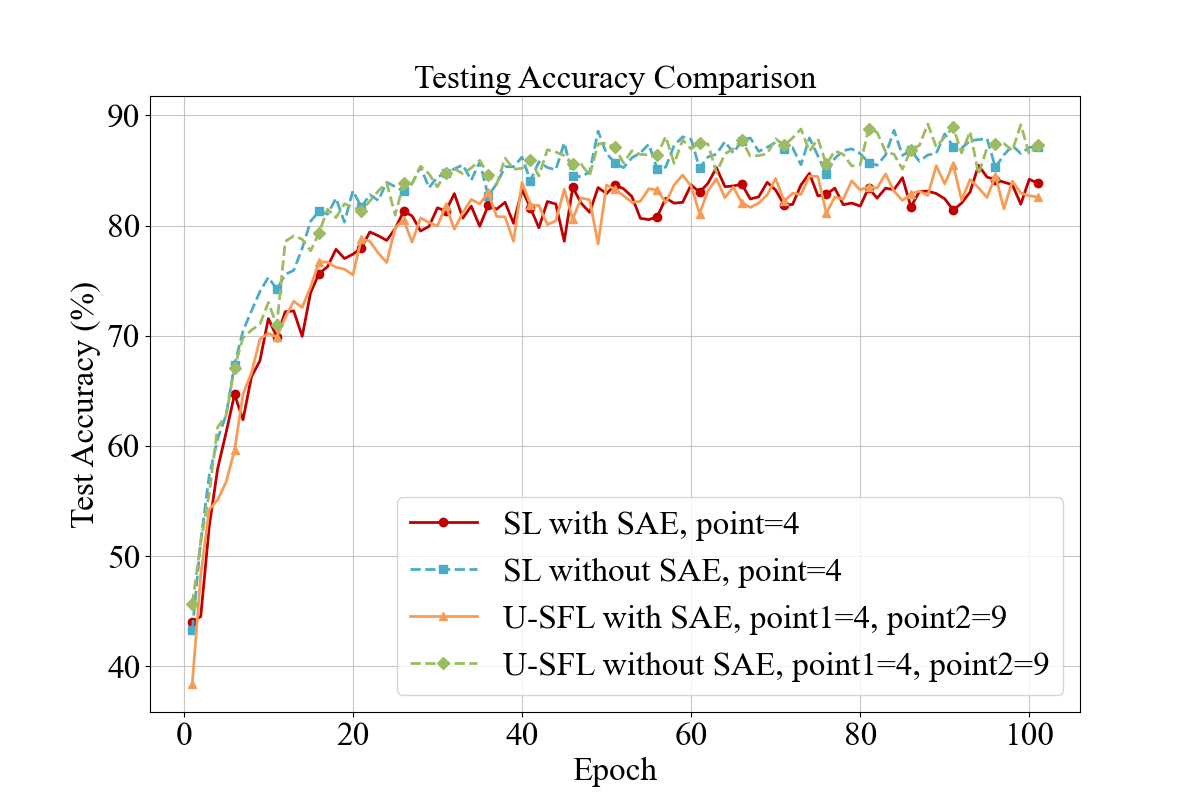}
            \label{fig:Caltech-101_withSAE_withoutSAE_testing}
        } &
        \subfigure[Different split point combinations (Caltech-101, Testing)]{
            \includegraphics[width=0.26\textwidth]{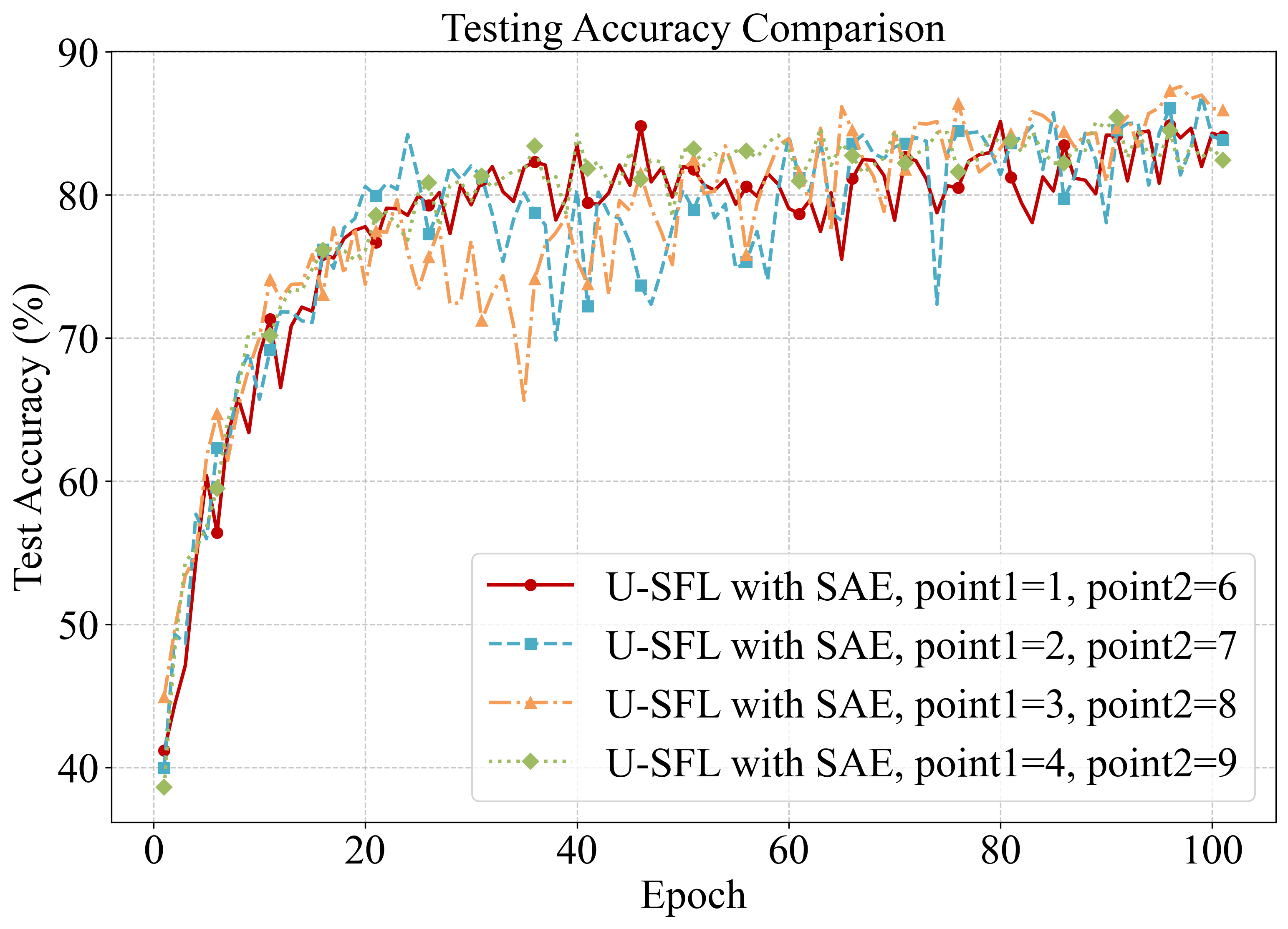}
            \label{fig:Caltech-101_DifferentPoints_testing}
        } \\
        \subfigure[CIFAR-10 (Training)]{
            \includegraphics[width=0.3\textwidth]{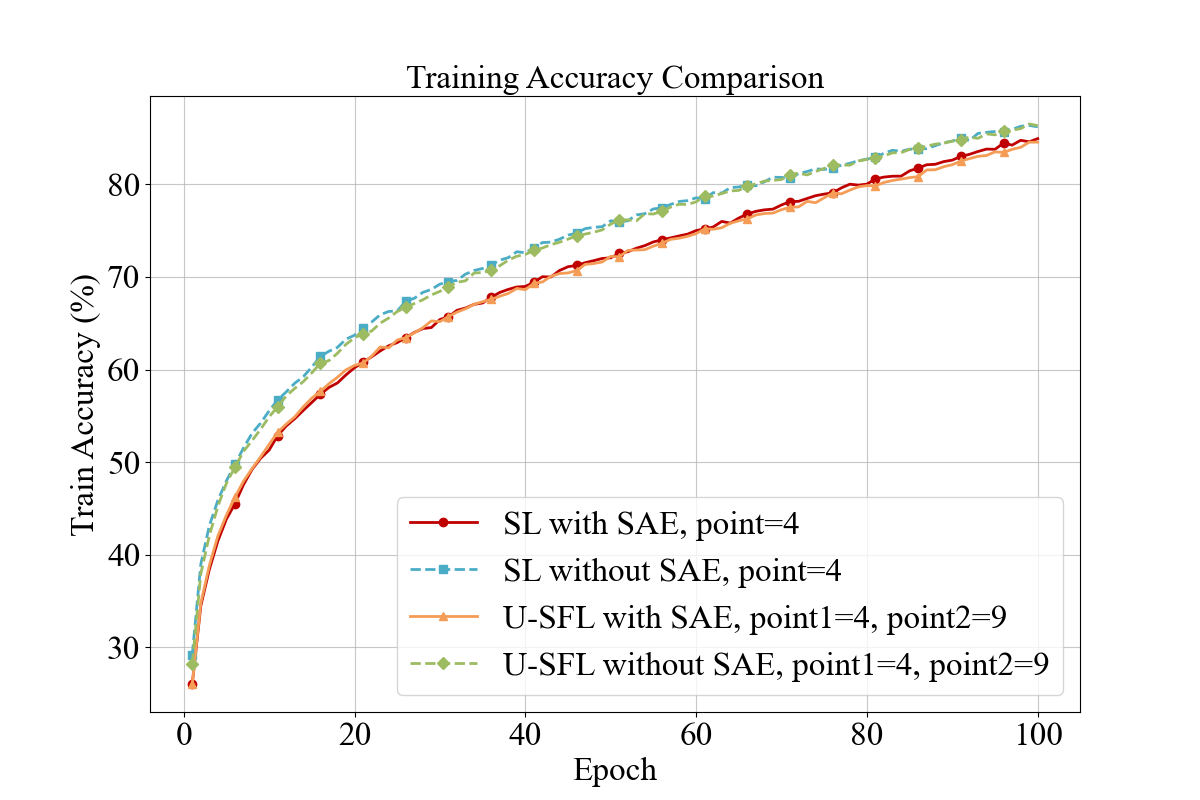}
            \label{fig:CIFAR-10_withSAE_withoutSAE_training}
        } &
        \subfigure[CIFAR-10 (Testing)]{
            \includegraphics[width=0.3\textwidth]{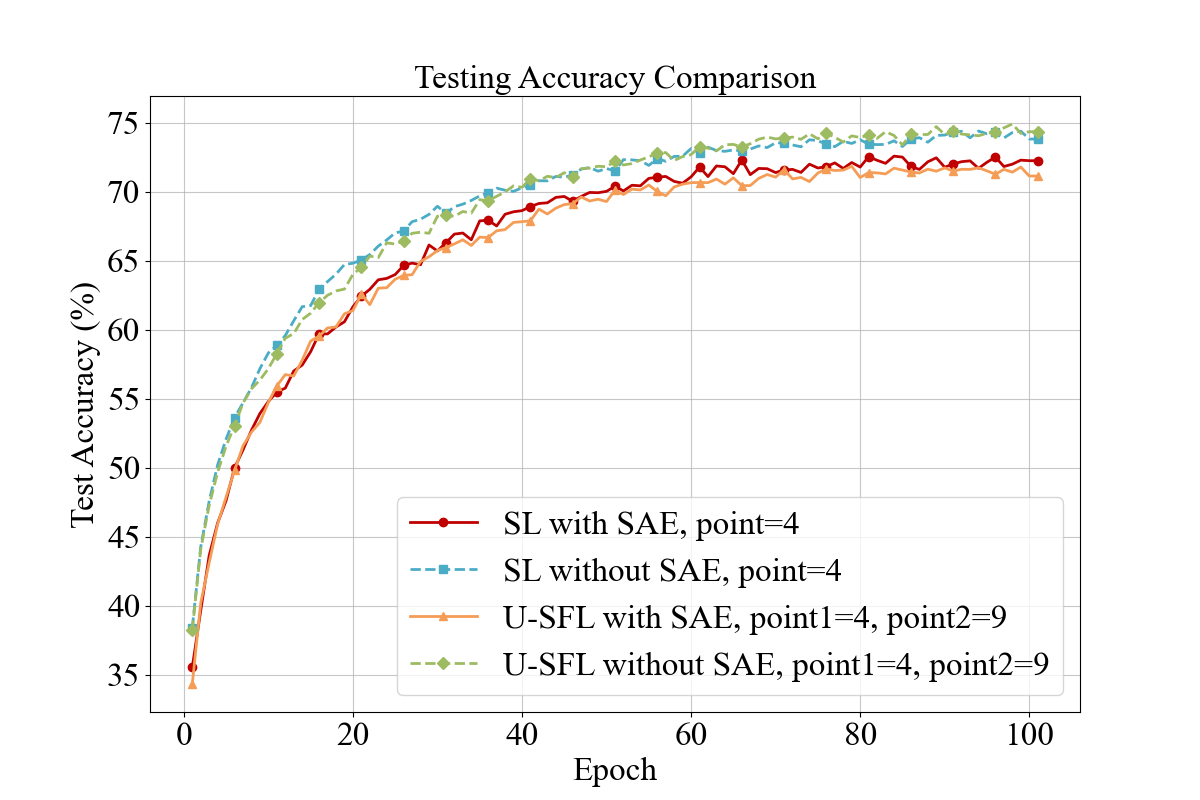}
            \label{fig:CIFAR-10_withSAE_withoutSAE_testing}
        } &
        \subfigure[Different split point combinations (CIFAR-10, Testing)]{
            \includegraphics[width=0.26\textwidth]{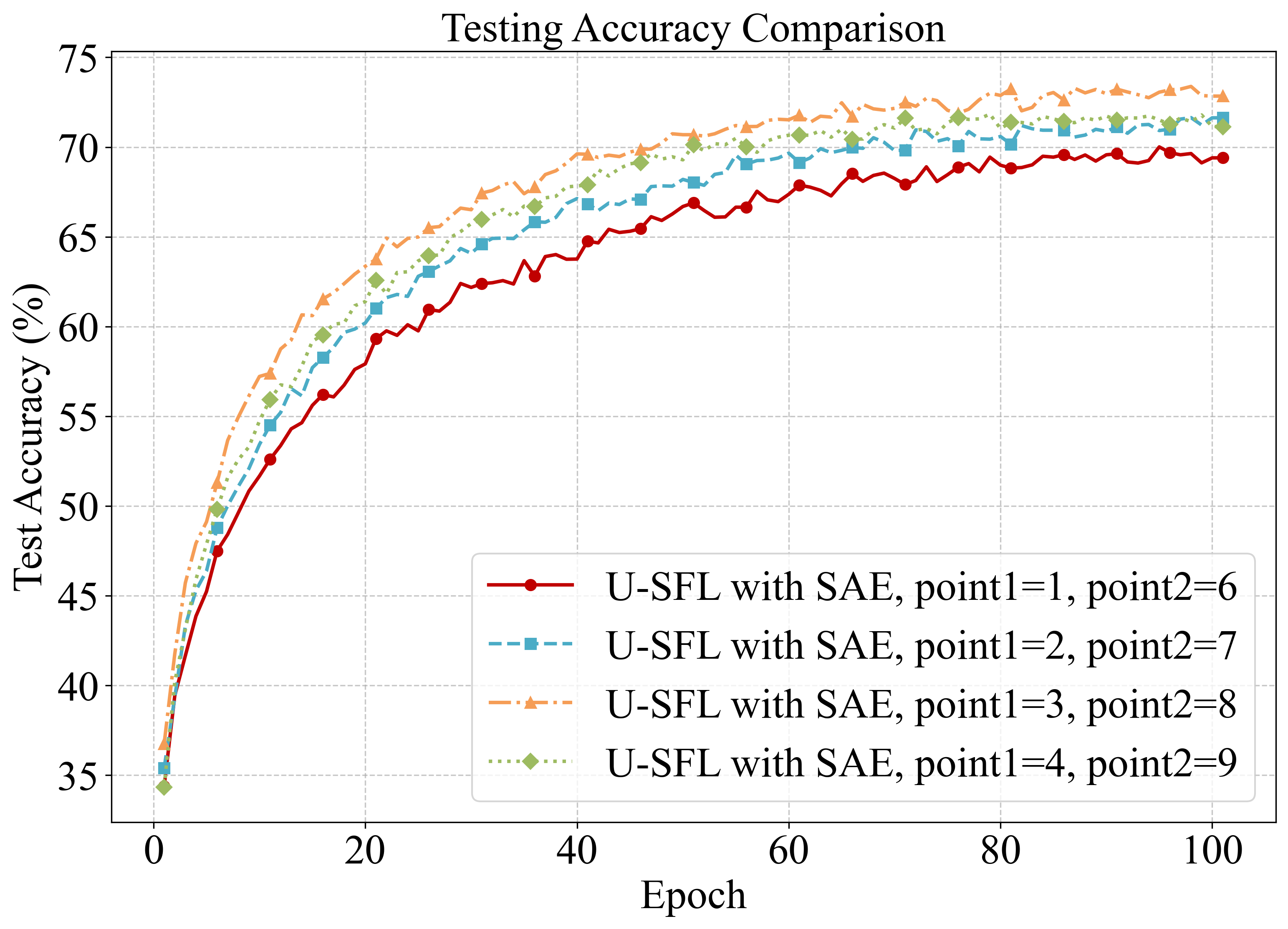}
            \label{fig:CIFAR-10_DifferentPoints_testing}
        }
    \end{tabular}
    \caption{Training and testing results of different model configurations with different datasets ((a) and (d) for training accuracy, (b) and (e) for testing accuracy). Testing accuracy comparison for different split point combinations ((c) and (f)).}
    \label{fig:Convergence_Performance}
\end{figure*}


\subsubsection{Semantic-aware Auto-Encoder (SAE) architecture}
The semantic-aware auto-encoder (SAE) architecture, inspired by the deep JSCC scheme \cite{JSCC}, consists of an encoder and a decoder, both utilizing convolutional layers with PReLU activations. Table \ref{table:sae_structure} summarizes the key components of the SAE structure.

\begin{table}[h]
\centering
\caption{The structure of the SAE in the proposed U-SFL framework}
\label{table:sae_structure}
\begin{tabular}{|c|>{\centering\arraybackslash}m{5cm}|}
\hline
\textbf{Component} & \textbf{LayerName}  \\
\hline
\multirow{7}{*}{\centering Encoder} 
 & Input Normalization \\
 & Conv+PReLU  \\
 & Conv+PReLU  \\
 & Conv+PReLU  \\
 & Conv+PReLU  \\
 & Conv+PReLU \\
 & Output Normalization  \\
\hline
\multirow{7}{*}{\centering Decoder}
 & Input Normalization  \\
 & TransConv+PReLU  \\
 & TransConv+PReLU  \\
 & TransConv+PReLU  \\
 & TransConv+PReLU  \\
 & TransConv+Sigmoid  \\
 & Output Denormalization  \\
\hline
\end{tabular}
\end{table}

\subsubsection{Split Point Selection}
We use the ResNet-18 architecture as the base model. The model structure and partition points of the ResNet-18 are shown in Fig. \ref{fig:ResNet-18}.
In our U-SFL framework, we carefully selected the split point combinations to balance computational load, communication efficiency, and model performance. The chosen combinations are (1,6), (1,7), (1,8), (1,9), (2,7), (2,8), (2,9), (3,8), (3,9), and (4,9), where the first number represents the split point between the VU in part a and the server in part b, and the second number represents the split point between the server in part b and the VU in part c.
These combinations were selected based on the following considerations. 
We ensure that the first split point is always in the earlier layers (1-4) to leverage the VU's computational capabilities while minimizing initial data transfer.
The second split point is chosen from later layers (6-9) to allow significant computation on the server side, which typically has more powerful resources.
We avoid extreme splits (e.g., (5,6)) that could lead to significant imbalances in computation or excessive communication overhead.

\subsubsection{Model Configurations}
For both datasets, we compared four distinct model configurations: traditional SL with partition at layer 4 (SL without SAE), traditional SL with partition at layer 4 with SAE (SL with SAE), U-SFL with partitions at layers 4 and 9 (U-SFL without SAE), U-SFL with partitions at layers 4 and 9 with SAE (U-SFL with SAE). To analyze the influence of different split point combinations on the classification performance of U-SFL with SAE, we evaluated four split point combinations: (1, 6), (2, 7), (3, 8) and (4, 9).

\subsubsection{Convergence Performance Comparison}
Fig. \ref{fig:Caltech-101_withSAE_withoutSAE_training} and Fig. \ref{fig:CIFAR-10_withSAE_withoutSAE_training} illustrate the train accuracy curves for all four configurations over the training for Caltech-101 and CIFAR-10, respectively. 
Fig. \ref{fig:Caltech-101_withSAE_withoutSAE_testing} and Fig. \ref{fig:CIFAR-10_withSAE_withoutSAE_testing} illustrate test accuracy curves for all four configurations over the training for Caltech-101 and CIFAR-10, respectively.
The proposed U-SFL method demonstrates comparable performance to traditional SL across both datasets. This equivalence in classification accuracy is maintained despite U-SFL's more complex architecture involving three-part splitting of the model. These results strongly support the viability of U-SFL as an effective approach for implementing distributed SL without significant performance degradation. The U-SFL offers potential advantages in terms of parallel processing and enhanced privacy protection through its unique model partitioning strategy.

The integration of SAE into both SL and U-SFL configurations results in a marginal decrease in accuracy for both Caltech-101 and CIFAR-10 datasets.  But this reduction is minimal and does not substantially impact the overall performance of either SL or U-SFL methods. SAE may contribute to data privacy by encoding feature maps, potentially obfuscating original inputs. SAE's dimension reduction capability could decrease data transfer between split points. Given these potential advantages, the observed minor decrease in accuracy can be viewed as an acceptable trade-off. 

Fig. \ref{fig:Caltech-101_DifferentPoints_testing} and Fig. \ref{fig:CIFAR-10_DifferentPoints_testing} show the testing accuracy comparison for different split point combinations on Caltech-101 (top) and CIFAR-10 (bottom) datasets. These results suggest that the choice of split points can significantly impact the model's performance, especially on more challenging datasets like CIFAR-10. Generally, splitting at later layers (e.g., (3,8) and (4,9)) tends to yield better performance, possibly due to more comprehensive feature extraction on the VU side before the first split, a better balance of computation between the VU and server sides and reduced information loss during the splitting process. However, the optimal split point combination may vary depending on the specific dataset and task. For Caltech-101, the performance differences are less significant, suggesting that the model is more robust to split point selection for this dataset.
These findings highlight the importance of carefully selecting split points in U-SFL to optimize performance.

\subsection{Communication Efficiency Analysis}
To evaluate the communication efficiency of our proposed U-SFL framework with and without the SAE, we conducted experiments to compare the data transmission sizes at different split points. 
The CIFAR-10 dataset with input image dimensions of 32x32x3 was adopted in our experimental design.
Since the SAE component is placed between client-side a and server-side b, we placed the SAE at cut layer1. We analyzed four potential positions for cut layer1, corresponding to the outputs of layers 1, 2, 3, and 4 of the ResNet-18 model. When enabled, the SAE was designed to compress the feature maps by a factor of 4 in spatial dimensions and expand the channel dimension by a factor of 32. The experiment calculated the size of the smashed data (intermediate feature maps) at each potential split point, both with and without the SAE.

\begin{figure}[t]
    \centering
    \renewcommand{\figurename}{Fig.}
    \includegraphics[width=6cm]{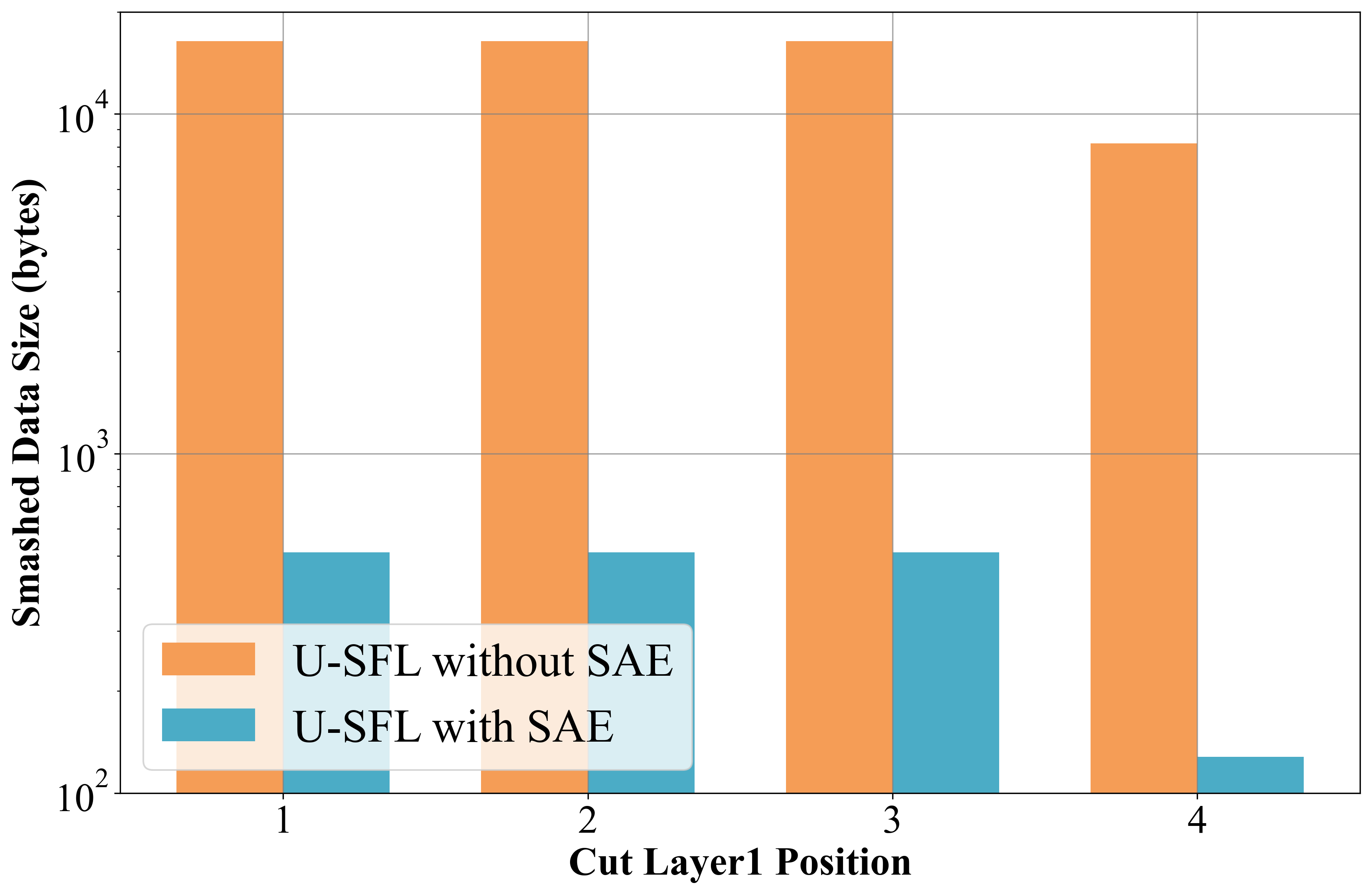}
    \caption{Comparison of smashed data size with and without SAE.}
    \label{fig:smashed_data_comparison}
\end{figure}

Fig. \ref{fig:smashed_data_comparison} illustrates the comparison of smashed data sizes with and without SAE at different cut layer positions. From the results, we can observe the SAE significantly reduces the size of the smashed data across all split points. 
The SAE maintains its efficiency across different split points, showing a consistent reduction in data size.
As we move to deeper layers (from cut layer 1 to 4), the smashed data size without SAE decreases due to the natural dimensionality reduction in the network. However, the SAE still provides substantial benefits, especially in the earlier layers where the feature maps are larger.
While the SAE significantly reduces data transmission, it's important to note that this comes at the cost of additional computation on the VU side. This trade-off between communication efficiency and computational overhead is a key consideration in the U-SFL framework.                         
 
\begin{figure}[t]
    \centering
    \renewcommand{\figurename}{Fig.}
    \includegraphics[width=6cm]{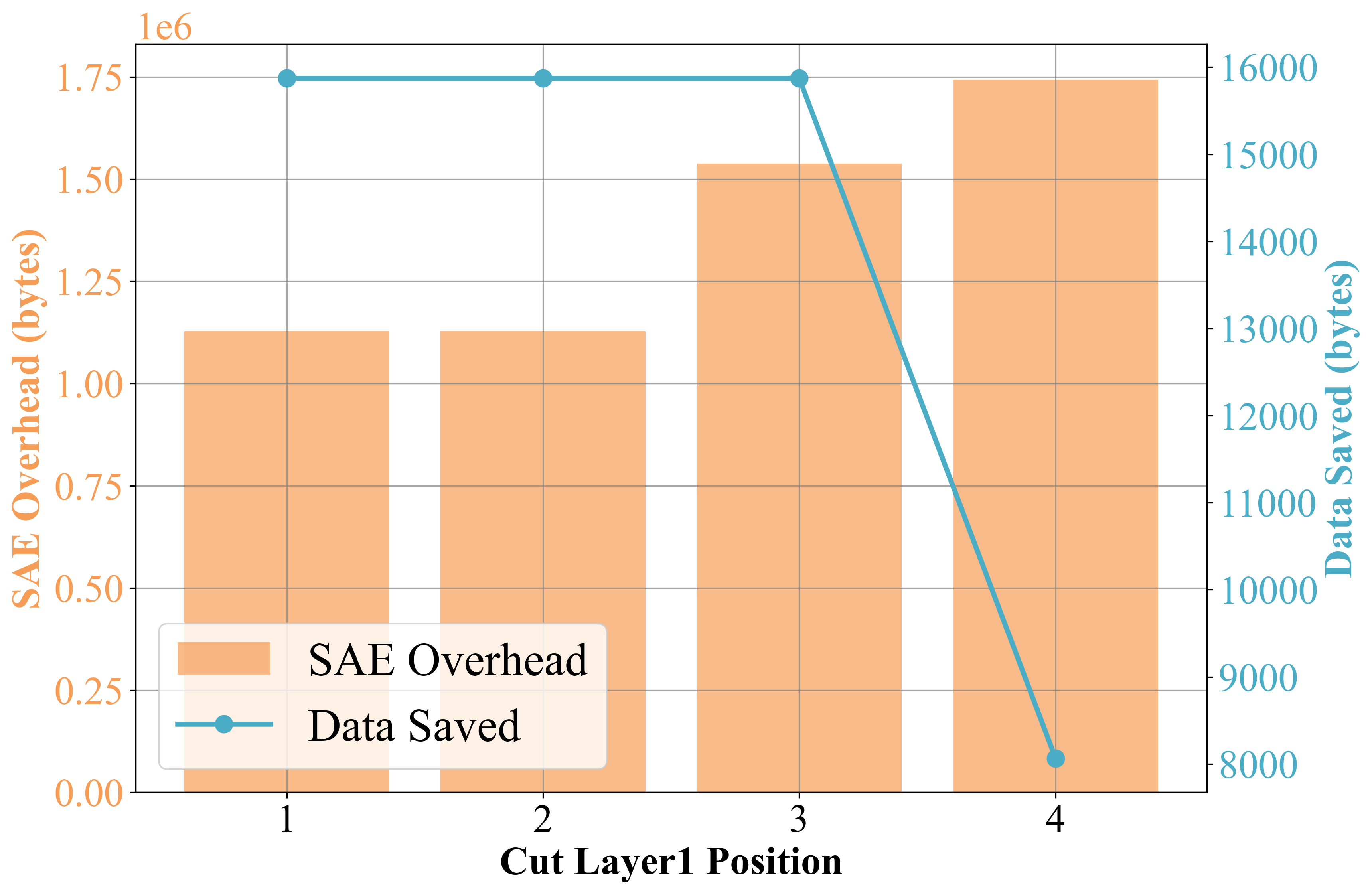}
    \caption{SAE overhead vs. data savings for different cut layers.}
    \label{fig:sae_overhead_vs_savings}
\end{figure}

To further investigate the trade-off between computational overhead and communication efficiency introduced by the SAE, we conducted an additional analysis. 
Fig. \ref{fig:sae_overhead_vs_savings} presents a dual-axis chart comparing SAE overhead (left y-axis, bars) with data savings (right y-axis, line) across different cut layers.
SAE overhead remains stable (1.1-1.2 million bytes) for cut layers 1-3, but increases significantly (1.75 million bytes) for layer 4.
Data savings are constant (16,000 bytes) for layers 1-3, but halve for layer 4. Cut layers 1-3 offer the optimal balance between overhead and savings. Layer 4 shows diminishing returns, with increased overhead and decreased savings.
These findings emphasize the importance of cut layer selection in the U-SFL framework. For resource-constrained vehicular Network environment, earlier cut layers (1-3) appear to offer a more favorable trade-off between computational load and communication efficiency. 

To further evaluate the efficiency of our proposed U-SFL framework with and without SAE, we conducted a simulation study on the communication and computation costs in a vehicular network scenario. We simulated various numbers of vehicles (from 5 to 30) and analyzed their performance under different network conditions. The simulation considers factors such as bandwidth allocation, transmission power, and network topology to provide a comprehensive analysis of the system's performance. The key components of our setup include: vehicle movement simulation with realistic parameters (speed, initial position, stay time); dynamic calculation of data rates based on distance from the edge server; computation of communication and computation overheads for different network splits; comparison of scenarios with and without SAE implementation.

\begin{figure}[t]
    \centering
    \renewcommand{\figurename}{Fig.}
    \includegraphics[width=6cm]{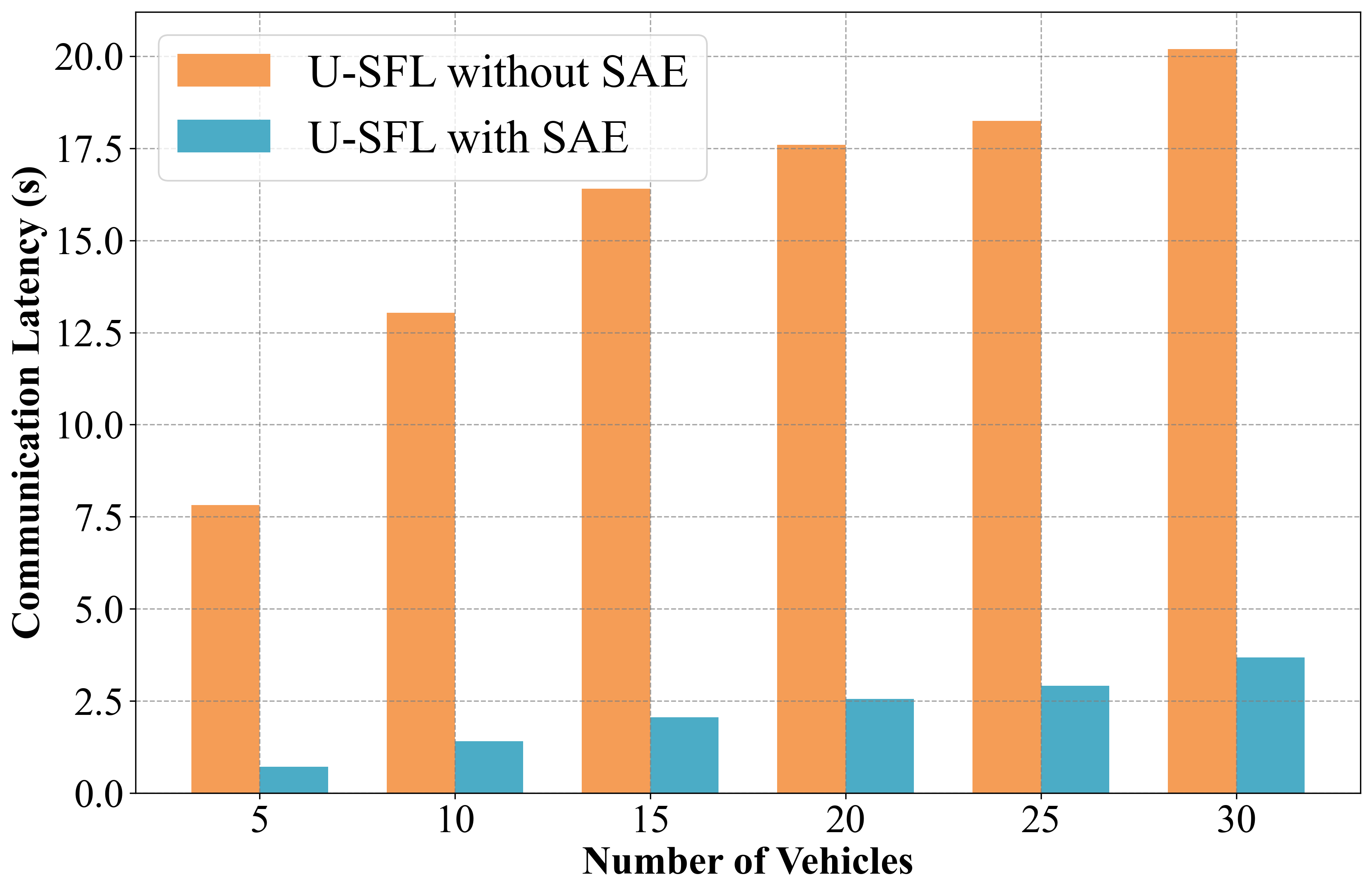}
    \caption{Communication latency comparison.}
    \label{fig:communication_latency}
\end{figure}

\begin{figure}[t]
    \centering
    \renewcommand{\figurename}{Fig.}
    \includegraphics[width=6cm]{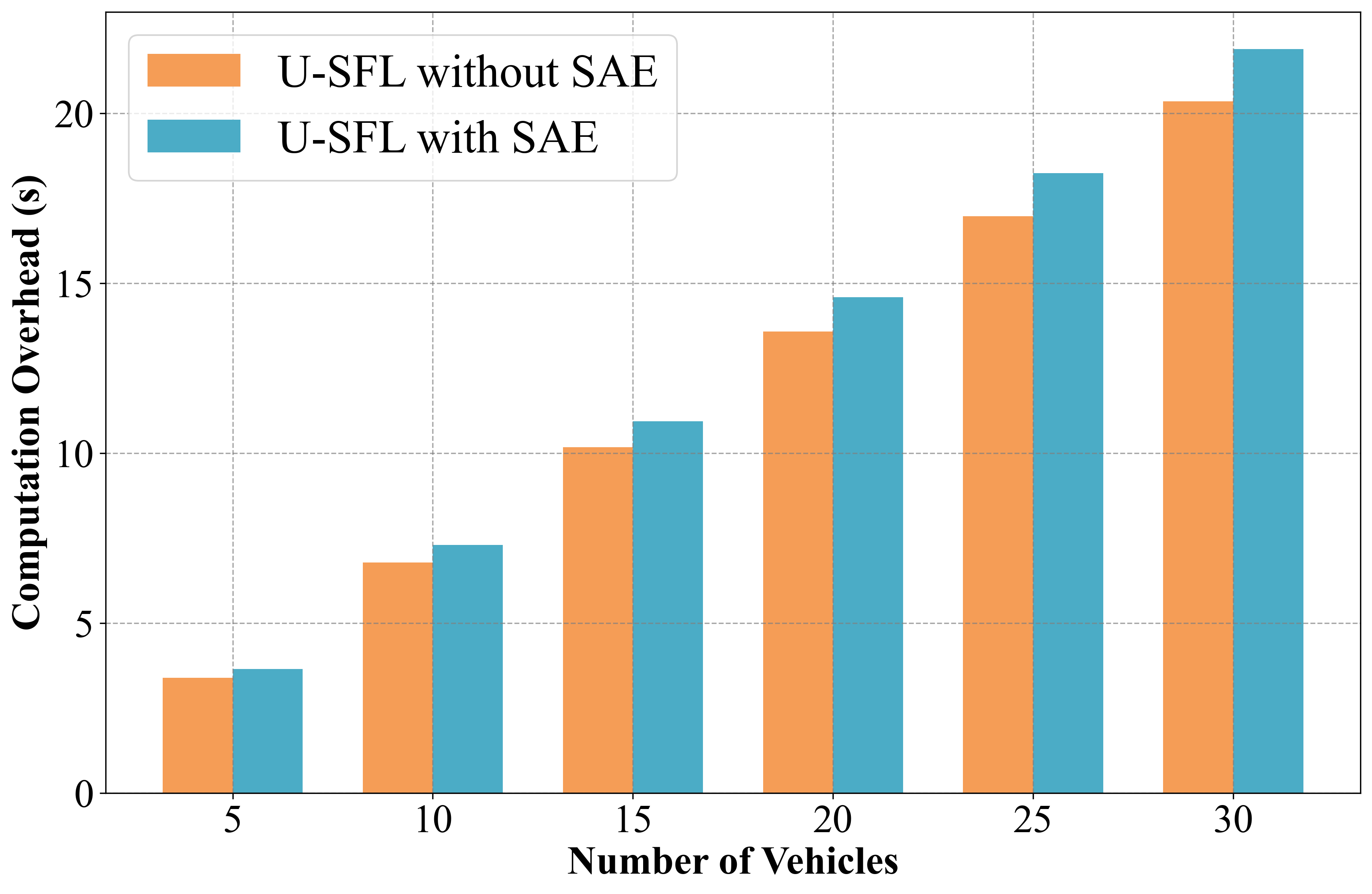}
    \caption{Computation latency comparison.}
    \label{fig:computation_latency}
\end{figure}

Fig. \ref{fig:communication_latency} and Fig. \ref{fig:computation_latency} illustrate the communication and computation latencies, respectively, for different numbers of vehicles in the network. Fig. \ref{fig:communication_latency} demonstrates that the U-SFL framework with SAE significantly reduces communication latency across all vehicle densities. The reduction is particularly pronounced as the number of vehicles increases, indicating SAE's effectiveness in mitigating network congestion.
Fig. \ref{fig:computation_latency} shows a slight increase in computation latency when SAE is employed. This is expected due to the additional processing required by the SAE. However, the increase is relatively small compared to the significant reduction in communication latency.
The results indicate that SAE effectively redistributes the network load, shifting some of the burden from communication to computation. This is particularly beneficial in vehicular networks where communication resources are often more constrained than computational resources.

\begin{figure}[t]
    \centering
    \renewcommand{\figurename}{Fig.}
    \includegraphics[width=6cm]{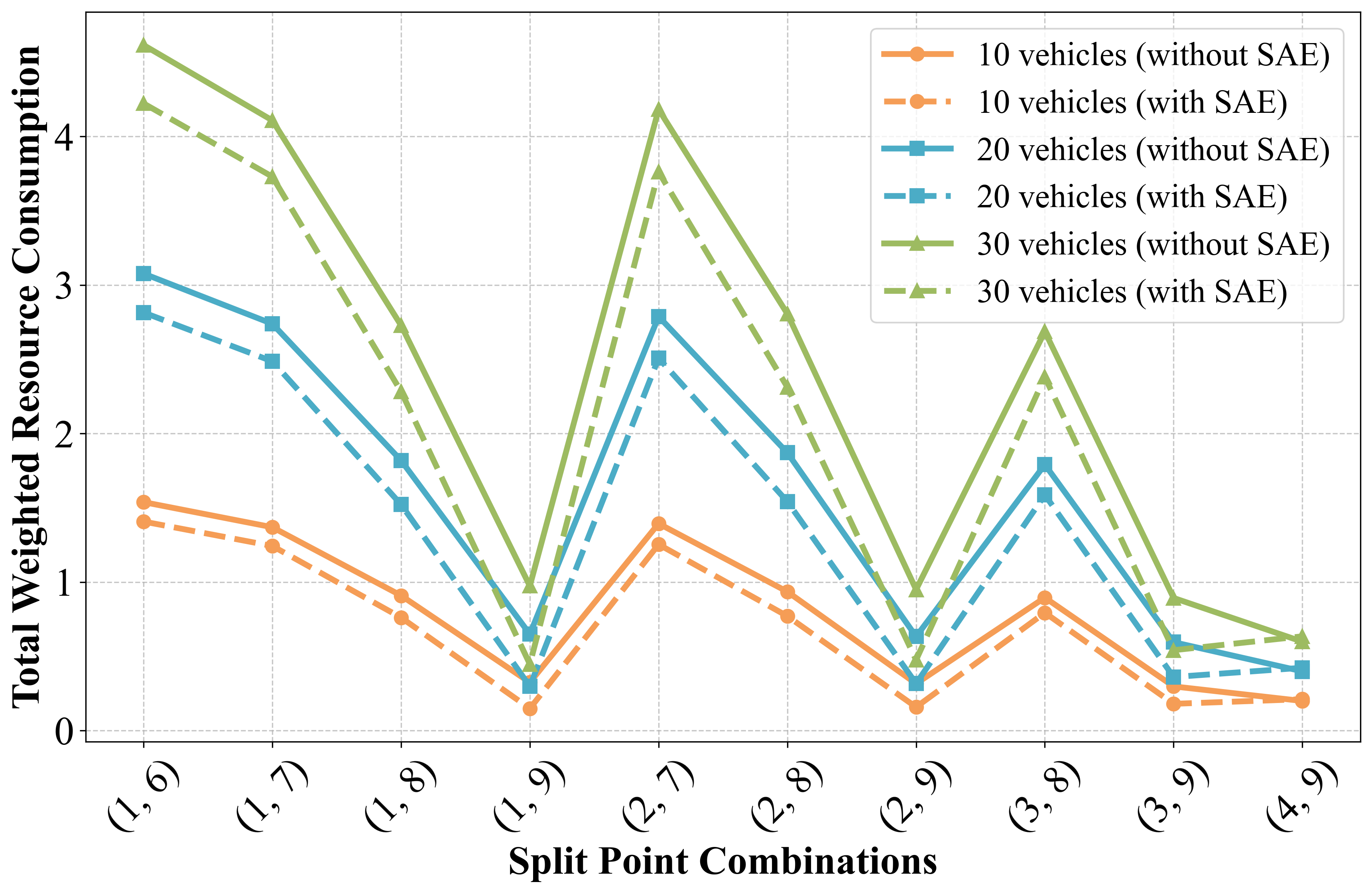}
    \caption{Comparison of weighted resource consumption across different scenarios.}
    \label{fig:weighted_resource_consumption}
\end{figure}

To comprehensively evaluate the performance of our U-SFL framework, we conducted an experiment to analyze the communication efficiency across different split point combinations. We define communication efficiency as a weighted sum of latency and energy consumption, with weights of 0.7 and 0.3 respectively, reflecting the greater importance of latency in vehicular networks.
Fig. \ref{fig:weighted_resource_consumption} illustrates the comparison of weighted resource consumption across different scenarios and split point combinations. Across all vehicle densities and split point combinations, the implementation of SAE consistently reduces the total weighted resource consumption. This reduction is particularly significant in scenarios with higher vehicle counts, demonstrating the scalability of the SAE approach. The results also indicate that split point combinations (1,9) and (2,9) generally yield the lowest resource consumption across all scenarios. Split points that occur earlier in the network (e.g., (1,6), (1,7)) generally show higher resource consumption compared to later splits. This trend is consistent across all vehicle densities, suggesting that optimizing the split point location is crucial for system efficiency.

\subsection{DRL Based Multi-objective Convergence Performance}
$\textit{Environment.}$ 
In our experiment, we consider a vehicular edge network with $I = 5$ VUs and an ES. The distance between each VU and the ES is uniformly distributed as $d_i \sim U[10, 100]$ meters. At the beginning of each episode, each VU receives $H_i \sim Pois(\lambda_p)$ tasks, where $Pois(\lambda_p)$ denotes the Poisson distribution with parameter $\lambda_p$ set to 100 for training and 5 for testing. An episode terminates when all tasks are completed or when the maximum number of steps (50) is reached. The channel gain is calculated using a path loss model: $20 \log_{10}(d_i) + 20 \log_{10}(2.4\times10^9) - 147.55$ \cite{GANPowered}, where $d_i$ is the distance in meters. We consider a dynamic channel setting where the maximum bandwidth for each VU is 10 MHz, and the maximum computing frequency is 2.5 GHz. The background noise power is set to $10^{-10}$ W.


$\textit{Agent.}$ 
Each actor network consists of a base module followed by four output heads. The base module is composed of fully connected layers with two hidden layers of 256 and 128 neurons respectively, LeakyReLU activation functions, and two residual blocks for improved gradient flow. The output of the base module is then passed through an attention mechanism before being fed into four separate output heads, each corresponding to a different action component: two for partition points selection, one for bandwidth allocation, and one for frequency allocation.
The critic network has a similar structure to the actor's base module, with an additional output layer of a single neuron to estimate the state value.

We train the actors and the critic using Adam optimizer with an initial learning rate of $10^{-4}$ for both, and a learning rate decay factor of 0.9999. The discount factor $\gamma$ is set to 0.99, and the GAE parameter $\lambda$ is 0.95. The PPO clip range $\epsilon$ is set to 0.1, and the entropy coefficient $\zeta$ is 0.005.
The training process consists of 30,000 episodes, with a maximum of 50 steps per episode. After each episode, we perform multiple updates (epochs) on the neural networks using the collected data. The number of update epochs after each episode is calculated as $J \times (\|\mathcal{M}\| / B)$, where the size of the experience replay buffer $\|\mathcal{M}\|$ is 1024, the batch size $B$ is 256, and the sample reuse time $J$ is 5.  
We implement an $\epsilon$-greedy strategy for exploration, with $\epsilon$ starting at 0.1 and decaying over time. To enhance stability during training, we employ gradient clipping with a maximum norm of 1.0 for both actor and critic networks. We also use a state normalizer to normalize the input states, which helps in stabilizing the learning process.


\begin{figure}[htbp]
    \centering
    \renewcommand{\figurename}{Fig.}
    \subfigure[Latency]{
        \includegraphics[width=0.3\textwidth]{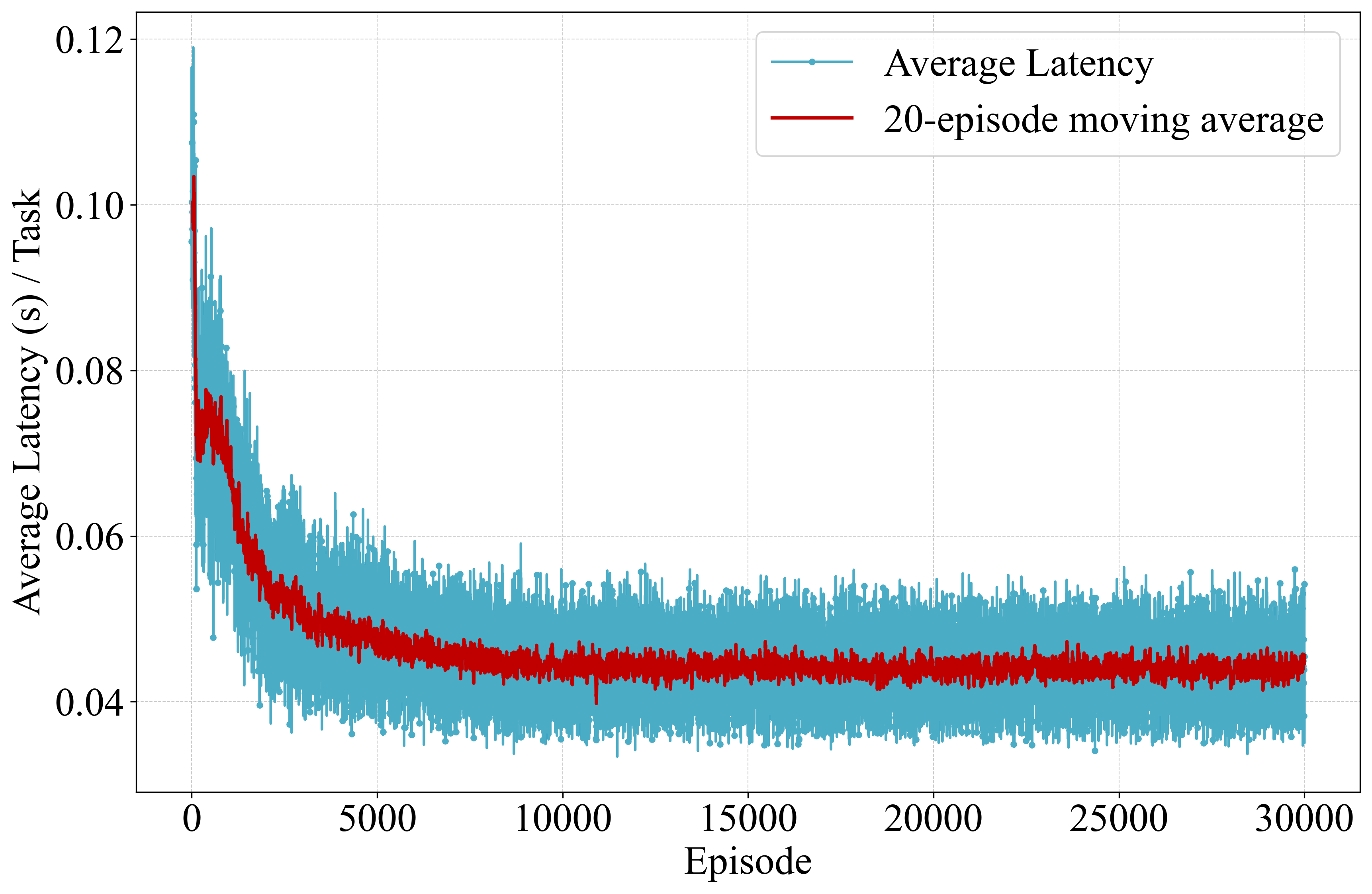}
        \label{fig:average_latency}
    }
    \subfigure[Energy consumption]{
        \includegraphics[width=0.3\textwidth]{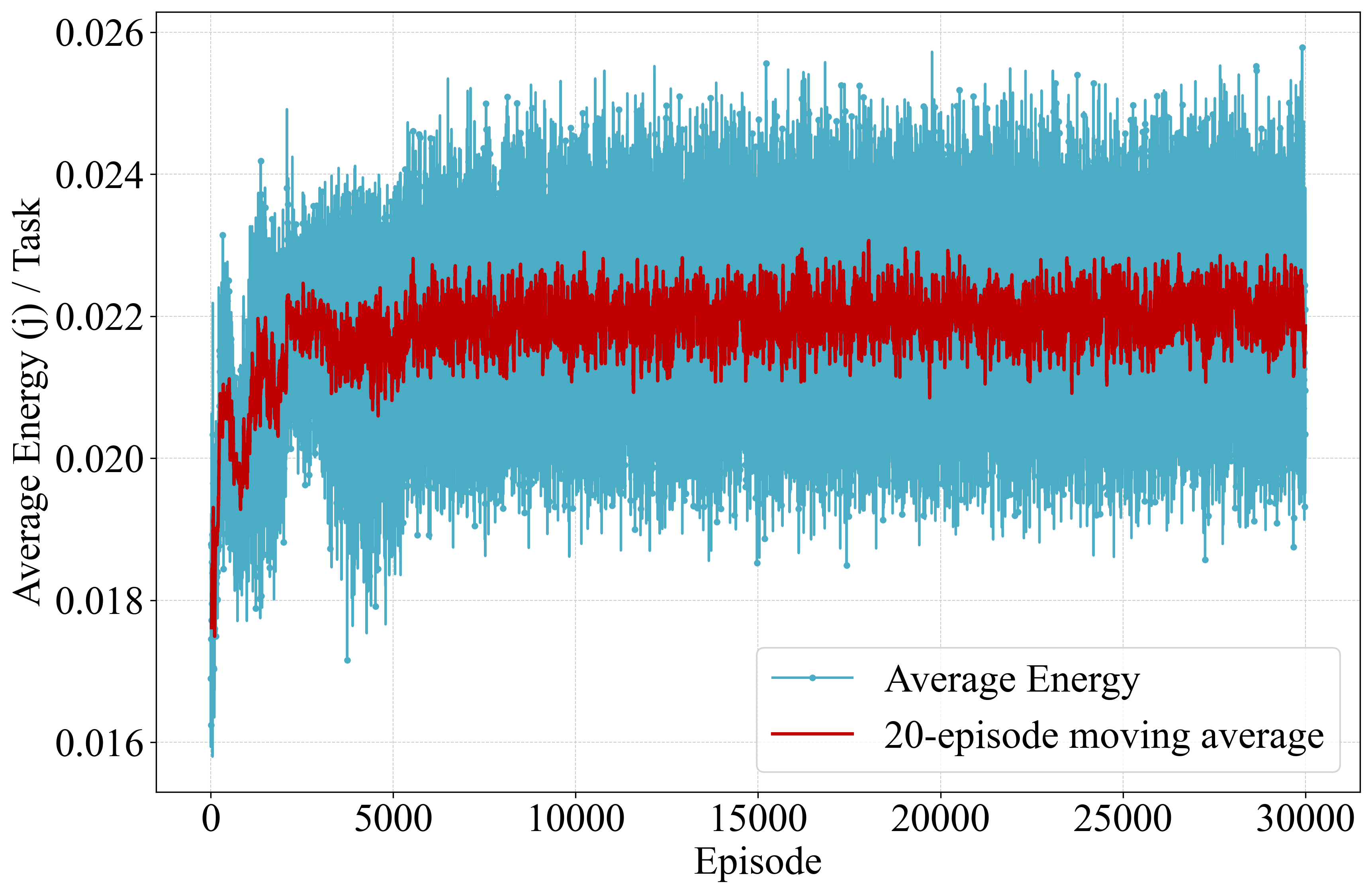}
        \label{fig:average_energy}
    }
    \caption{Convergence performance of our proposed DRL based multi-objective optimization algorithm.}
    \label{fig:Convergence Performance}
\end{figure}

$\textit{Results and Analysis.}$ 
Fig. \ref{fig:Convergence Performance} illustrates the convergence performance of our proposed DRL algorithm.
We analyze two key metrics: average latency and average energy consumption.
As shown in Fig. \ref{fig:average_latency}, 
the 20-episode moving average (redline) demonstrates a consistent downward trend, eventually stabilizing at around 0.045 s after 15,000 episodes. This indicates that our algorithm successfully optimizes the task offloading and resource allocation to minimize processing delays.
As shown in Fig. \ref{fig:average_energy}, 
the 20-episode moving average (redline) indicates a stable trend with minor fluctuations, suggesting that the algorithm effectively balances energy efficiency with other objectives.   
The convergence patterns observed across the two metrics demonstrate the effectiveness of our proposed approach in addressing the multi-objective optimization problem in vehicular edge computing networks. The algorithm shows rapid initial learning and long-term stability, indicating its potential for practical applications in dynamic vehicular networks.

\section{Conclusion}
In this paper, we proposed a novel U-SFL framework for vehicular edge networks, integrating a SAE to optimize communication efficiency.
The U-SFL framework demonstrates comparable classification performance to traditional SL while offering enhanced privacy protection and parallel processing capabilities. 
Our results underscore the crucial role of split point selection in model performance. 
The U-SFL framework with SAE substantially reduces data transmission volume and communication latency, especially as the number of vehicles increases. 
Our proposed DRL-based multi-objective optimization algorithm demonstrates good convergence performance in balancing latency, energy consumption, and cumulative reward. 
These findings collectively demonstrate the efficacy of our U-SFL framework in enhancing communication efficiency, preserving privacy, and optimizing resource utilization in vehicular edge networks. 
Future research directions could explore dynamic split point selection mechanisms, adaptation to more complex network topologies, and integration with emerging 6G technologies.


\begin{thebibliography}{1}
\bibitem{autonomousdriving_review} X. Ge, ``Ultra-Reliable Low-Latency Communications in Autonomous Vehicular Networks," in \textit{IEEE Transactions on Vehicular Technology}, vol. 68, no. 5, pp. 5005--5016, May 2019.

\bibitem{ad_benefits} S. Mozaffari, O. Y. Al-Jarrah, M. Dianati, P. Jennings and A. Mouzakitis, ``Deep Learning-Based Vehicle Behavior Prediction for Autonomous Driving Applications: A Review," in \textit{IEEE Transactions on Intelligent Transportation Systems}, vol. 23, no. 1, pp. 33--47, Jan. 2022.

\bibitem{v2x_overview} K. Sehla, T. M. T. Nguyen, G. Pujolle and P. B. Velloso, ``Resource Allocation Modes in C-V2X: From LTE-V2X to 5G-V2X," in \textit{IEEE Internet of Things Journal}, vol. 9, no. 11, pp. 8291--8314, 1 June1, 2022.

\bibitem{ad_vn_challenges} L. Chen et al., ``Milestones in autonomous driving and intelligent vehicles: Survey of surveys," in \textit{IEEE Trans. Intell. Veh.}, vol. 8, no. 2, pp. 1046--1056, Sep. 2023.


\bibitem{privacy_concerns} C. Xu, H. Wu, H. Liu, W. Gu,Y. Li, and D. Cao, ``Blockchain-oriented privacy protection of sensitive data in the internet of vehicles," in \textit{IEEE Trans. Intell. Veh.}, vol. 8, no. 2, pp. 1057--1067, Feb. 2023.


\bibitem{VN_communication_challenges} S. Moon and Y. Lim, ``Split and federated learning with mobility in vehicular edge computing," in \textit{2023 IEEE/ACIS 21st International Conference on Software Engineering Research, Management and Applications (SERA)}. IEEE, 2023, pp. 35--38.

\bibitem{vehicle_resource_constraints} X. Zhang, J. Liu, T. Hu, Z. Chang, Y. Zhang, and G. Min, ``Federated
learning-assisted vehicular edge computing: Architecture and research directions," in \textit{IEEE Vehicular Technology Magazine}, pp. 2--11, 2023.

\bibitem{fl_overview}  B. McMahan, E. Moore, D. Ramage, S. Hampson, and B. A. y Arcas, ``Communication-efficient Learning of Deep Networks From Decentralized Data," in \textit{Proc. AISTATS}, Apr. 2017.

\bibitem{fl_comm_challenge} M. F. Pervej, R. Jin, and H. Dai, ``Resource constrained vehicular edge federated learning with highly mobile connected vehicles," in \textit{IEEE Journal on Selected Areas in Communications}, 2023.

\bibitem{fl_resource_limitation} H. Xiao, J. Zhao, Q. Pei, J. Feng, L. Liu, and W. Shi, ``Vehicle selection and resource optimization for federated learning in vehicular edge computing," in \textit{IEEE Transactions on Intelligent Transportation Systems}, vol. 23, no. 8, pp. 11 073--11 087, 2021.

\bibitem{sl_sequential_limitation} J. Lee, M. Seif, J. Cho and H. Vincent Poor, ``Exploring the Privacy-Energy Consumption Tradeoff for Split Federated Learning," in \textit{IEEE Network}.

\bibitem{sl_privacy_concern} Y. Wu, Y. Song, T. Wang, L. Qian, and T. Q. Quek, ``Non-orthogonal multiple access assisted federated learning via wireless power transfer: A cost-efficient approach," in \textit{IEEE Transactions on Communications}, vol. 70, no. 4, pp. 2853--2869, 2022.

\bibitem{sl_label_attack} M. Wu et al., ``Federated Split Learning With Data and Label Privacy Preservation in Vehicular Networks," in \textit{IEEE Transactions on Vehicular Technology}, vol. 73, no. 1, pp. 1223--1238, Jan. 2024.

\bibitem{sl_communication_challenge} M. Wu, R. Yang, X. Huang, Y. Wu, J. Kang and S. Xie, ``Joint Optimization of Model Partition and Resource Allocation for Split Federated Learning over Vehicular Edge Networks," in \textit{IEEE Transactions on Vehicular Technology}.

\bibitem{resource_allocation_np_hard} X. Chen, L. Jiao, W. Li, and X. Fu, ``Efficient multi-user computation offloading for mobile-edge cloud computing," in \textit{IEEE/ACM Trans. Netw.}, vol. 24, no. 5, pp. 2795--2808, Oct. 2016.







\bibitem{SL- first_proposed} O. Gupta and R. Raskar, ``Distributed learning of deep neural network over multiple agents," in \textit{Journal of Network and Computer Applications}, vol. 116, pp. 1--8, 2018. [Online]. 

\bibitem{SL-health} P. Vepakomma, O. Gupta, T. Swedish, and R. Raskar, ``Split Learning for Health: Distributed Deep Learning Without Sharing Raw Patient Data," \textit{ arXiv preprint arXiv:1812.00564}, Dec. 2018.

\bibitem{SL-PSL1} M. Kim, A. DeRieux, and W. Saad, ``A Bargaining Game for Personalized, Energy Efficient Split Learning over Wireless Networks," in \textit{Proc. WCNC}, Mar. 2023.

\bibitem{SL-PSL2} P. Joshi, C. Thapa, S. Camtepe, M. Hasanuzzamana, T. Scully, and H. Afli, ``Splitfed Learning Without Client-side Synchronization: Analyzing Client-side Split Network Portion Size to Overall Performance," \textit{arXiv preprint arXiv:2109.09246}, Sep. 2021.

\bibitem{SL-SFL1} Y. Gao, M. Kim, C. Thapa, A. Abuadbba, Z. Zhang, S. Camtepe, H. Kim, and S. Nepal, ``Evaluation and optimization of distributed machine learning techniques for internet of things," in \textit{IEEE Transactions on Computers}, vol. 71, no. 10, pp. 2538--2552, 2021.

\bibitem{SL-SFL2} M. Gawali, C. Arvind, S. Suryavanshi, H. Madaan, A. Gaikwad, K. Bhanu Prakash, V. Kulkarni, and A. Pant, ``Comparison of privacypreserving distributed deep learning methods in healthcare," in \textit{Annual Conference on Medical Image Understanding and Analysis}. Springer,
2021, pp. 457--471.

\bibitem{SL-SFL3} C. Thapa, P. C. M. Arachchige, S. Camtepe, and L. Sun, ``Splitfed: When federated learning meets split learning," in \textit{Proceedings of the AAAI Conference on Artificial Intelligence}, vol. 36, no. 8, 2022, pp. 8485--8493.

\bibitem{SL-aggregate1} V. Turina, Z. Zhang, F. Esposito, and I. Matta, ``Combining split and federated architectures for efficiency and privacy in deep learning," in \textit{Proceedings of the 16th International Conference on emerging Networking Experiments and Technologies}, 2020, pp. 562--563.

\bibitem{SL-aggregate2} X. Liu, Y. Deng, and T. Mahmoodi, ``Energy efficient user scheduling for hybrid split and federated learning in wireless uav networks," in \textit{IEEE International Conference on Communications (ICC)}, 2022, pp. 1--6.

\bibitem{SL-auto_encoder1} C.-Y. Hsieh, Y.-C. Chuang, and A.-Y. Wu, ``C3-SL: Circular Convolution-Based Batch-Wise Compression for Communication-Efficient Split Learning," in \textit{Proc. MLSP}, Aug. 2022.

\bibitem{SL-auto_encoder2} J. Shao and J. Zhang, ``Communication-computation Trade-off in Resource-constrained Edge Inference," in \textit{IEEE Commun. Mag}., vol. 58, no. 12, pp. 20--26, Dec. 2020.

\bibitem{SL-straggler} Z. Lin, G. Qu, X. Chen and K. Huang, ``Split Learning in 6G Edge Networks," in \textit{IEEE Wireless Communications}, vol. 31, no. 4, pp. 170--176, August 2024.

\bibitem{SL-cluster1} Z. Lin, G. Zhu, Y. Deng, X. Chen, Y. Gao, K. Huang, and Y. Fang, ``Efficient Parallel Split Learning over Resource-constrained Wireless Edge Networks," in \textit{arXiv preprint arXiv:2303.15991}, Mar. 2023.

\bibitem{SL-cluster2} W. Wu, M. Li, K. Qu, C. Zhou, X. Shen, W. Zhuang, X. Li, and W. Shi, ``Split Learning over Wireless Networks: Parallel Design and Resource Management," in \textit{IEEE J. Sel. Areas Commun.}, vol. 41, no. 4, pp. 1051--1066, Apr. 2023.

\bibitem{1} X. Zhang, Z. Qi, G. Min, W. Miao, Q. Fan and Z. Ma, ``Cooperative Edge Caching Based on Temporal Convolutional Networks," \textit{IEEE Transactions on Parallel and Distributed Systems}, vol. 33, no. 9, pp. 2093--2105, 2022.

\bibitem{crowdsensing} X. Li et al., ``Multi-View Matrix Factorization for Sparse Mobile Crowdsensing," in \textit{IEEE Internet of Things Journal}, vol. 9, no. 24, pp. 25767--25779, 15 Dec.15, 2022.

\bibitem{vehicular} P. Garrido et al., ``Optimizing the Transmission of Multimedia Content over Vehicular Networks," \textit{2022 International Conference on Computational Science and Computational Intelligence (CSCI)}, Las Vegas, NV, USA, 2022, pp. 1112--1116.

\bibitem{GANPowered} B. Yin, Z. Chen and M. Tao, ``Predictive GAN-Powered Multi-Objective Optimization for Hybrid Federated Split Learning," in \textit{IEEE Transactions on Communications}, vol. 71, no. 8, pp. 4544-4560, Aug. 2023.

\bibitem{SplitVehicular} M. Wu, R. Yang, X. Huang, Y. Wu, J. Kang and S. Xie, ``Joint Optimization of Model Partition and Resource Allocation for Split Federated Learning over Vehicular Edge Networks," in \textit{IEEE Transactions on Vehicular Technology}.

\bibitem{DRL} F. Jiang, K. Wang, L. Dong, C. Pan, and K. Yang, ``Stacked auto encoder based deep reinforcement learning for online resource scheduling in large-scale MEC networks," 2020, \textit{arXiv:2001.09223}.

\bibitem{GAE} J. Schulman, P. Moritz, S. Levine, M. I. Jordan, and P. Abbeel, ``High-dimensional continuous control using generalized advantage estimation," in \textit{Proc. Int. Conf. Learn. Representations}, 2016, pp. 1-14.

\bibitem{JSCC} E. Bourtsoulatze, D. B. Kurka and D. Gündüz, ``Deep Joint Source-channel Coding for Wireless Image Transmission," \textit{ICASSP 2019 - 2019 IEEE International Conference on Acoustics, Speech and Signal Processing (ICASSP)}, Brighton, UK, 2019, pp. 4774-4778. 

\end{thebibliography}
\end{document}